%% file: main.tex
\documentclass[10pt,twocolumn,letterpaper]{article}

\usepackage{cvpr}
\usepackage{times}
\usepackage{epsfig}
\usepackage{graphicx}
\usepackage{amsmath}
\usepackage{amssymb}
\usepackage{setspace}

\usepackage{inconsolata}

\usepackage{cite}

\usepackage{algorithm}
\usepackage{algpseudocode}

\usepackage{multirow}
\usepackage{rotating}
\usepackage{booktabs}

\usepackage{enumitem}
\usepackage[olditem,oldenum]{paralist}

\usepackage{alltt}
\usepackage{listings}

\usepackage{mysymbols}

\usepackage{url}
\usepackage{xspace}
\usepackage{comment}
\usepackage{color}
\usepackage{xcolor}
\usepackage{afterpage}
\usepackage{pdfpages}
\usepackage{framed}
\usepackage{fancybox}
\usepackage{cuted}
\usepackage{caption}
\usepackage{subcaption}
\usepackage{paralist}

\usepackage[pagebackref=true,breaklinks=true,letterpaper=true,colorlinks,bookmarks=false,citecolor=green]{hyperref}

\definecolor{orange}{rgb}{1,0.5,0}

\input definitions.tex

\newcommand{\sml}[1]{\textcolor{purple}{#1}}

\renewcommand{\xhdr}[1]{\vspace{2pt}\noindent\textbf{#1}}
\newcommand{\supp}[1]{\textbf{\textcolor{red}{#1}}}
\renewcommand{\supp}[1]{#1}

\newcommand{\newcontent}[1]{\textcolor{cyan}{\textbf{NEW}}~$[$#1$]$}
\renewcommand{\newcontent}[1]{#1}

\cvprfinalcopy %

\def\cvprPaperID{135} %

\input{space_saver.tex}

\ifcvprfinal\fi
\begin{document}

\title{Embodied Question Answering in \\ Photorealistic Environments with Point Cloud Perception}

\author{\input{authors.tex}}

\maketitle
\thispagestyle{empty}

\begin{abstract}
To help bridge the gap between \textbf{internet vision}-style problems
and the goal of  \textbf{vision for embodied perception}
we instantiate a large-scale navigation task --  Embodied Question Answering~\cite{embodiedqa} in photo-realistic environments (Matterport 3D). We thoroughly study navigation policies that utilize  3D \pointclouds, RGB images, or their combination. Our analysis of these models reveals several key findings.
We find that two seemingly naive navigation baselines, forward-only and random, are strong navigators
and challenging to outperform, due to the specific choice of
the evaluation setting presented by \cite{embodiedqa}.  We find a novel loss-weighting scheme we call Inflection Weighting to be important when training recurrent models for navigation with behavior cloning and are able to out perform the baselines with this technique. We find that \pointclouds provide a richer signal than RGB images for learning obstacle avoidance, motivating the use (and continued study) of 3D deep learning models for embodied navigation. %

\end{abstract}
\blfootnote{$^\dagger$ denotes equal contribution}

\input{sections/main/intro_v3}

\input{sections/main/related_work}

\input{sections/main/data}
\input{sections/main/approach}

\input{sections/main/results}

\input{sections/main/conclusion}

\begin{spacing}{0.8}
{\footnotesize
\xhdr{Acknowledgements.} This work was supported in part by NSF (Grant \# 1427300), AFRL, DARPA, Siemens, Samsung, Google, Amazon, ONR YIPs and ONR Grants N00014-16-1-\{2713,2793\}. The views and conclusions contained herein are those of the authors and should not be interpreted as necessarily representing the official policies or endorsements, either expressed or implied, of the U.S. Government, or any sponsor.
}
\end{spacing}

\newpage
{
    \small
    \bibliographystyle{ieeetr}
    \bibliography{strings,main}
}

\clearpage
\input{sections/supp/supp_content.tex}

\end{document}

%% file: definitions.tex
\newcommand{\myquote}[1]{\emph{`#1'}}
\newcommand{\myapprox}{{\raise.17ex\hbox{$\scriptstyle\sim$}}}
\newcommand{\xhdr}[1]{\vspace{3pt}\noindent\textbf{#1}}

\newcommand{\reffig}[1]{Fig.~\ref{#1}}

\newcommand{\reftab}[1]{Tab.~\ref{#1}}

\newcommand{\eqa}{EmbodiedQA\xspace}
\newcommand{\mpeqa}{MP3D-EQA\xspace}

\newcommand\blfootnote[1]{%
  \begingroup
  \renewcommand\thefootnote{}\footnote{#1}%
  \addtocounter{footnote}{-1}%
  \endgroup
}

\newcommand{\matterport}{Matterport3D\xspace}
\newcommand{\rgbd}{RGB-D\xspace}
\newcommand{\pointcloud}{point cloud\xspace}
\newcommand{\pointclouds}{point clouds\xspace}

%% file: space_saver.tex
\newlength{\sectionReduceTop}
\newlength{\sectionReduceBot}
\newlength{\subsectionReduceTop}
\newlength{\subsectionReduceBot}
\newlength{\abstractReduceTop}
\newlength{\abstractReduceBot}
\newlength{\captionReduceTop}
\newlength{\captionReduceBot}
\newlength{\subsubsectionReduceTop}
\newlength{\subsubsectionReduceBot}

\newlength{\eqnReduceTop}
\newlength{\eqnReduceBot}

\newlength{\horSkip}
\newlength{\verSkip}

\newlength{\figureHeight}

%% file: authors.tex
Erik Wijmans$^{1\dagger}$,
Samyak Datta$^{1\dagger}$, Oleksandr Maksymets$^{2\dagger}$,
Abhishek Das$^1$, \\ Georgia Gkioxari$^2$, Stefan Lee$^1$, Irfan Essa$^1$, Devi Parikh$^{1,2}$, Dhruv Batra$^{1,2}$
\\
$^1$Georgia Institute of Technology
~~
$^2$Facebook AI Research
\\
$^{1}$%
{\small \texttt{\{etw, samyak, abhshkdz, steflee, irfan, parikh, dbatra\}@gatech.edu}}\\
$^2$%
{\small \texttt{\{maksymets, gkioxari\}@fb.com}}

%% file: sections/main/intro_v3.tex
\vspace{\sectionReduceTop}
\section{Introduction}
\vspace{\sectionReduceBot}

Imagine asking a home robot \myquote{Hey - can you go check if my laptop is on my desk? And if so, bring it to me.} In order to be successful, such an agent would need a range of artificial intelligence (AI) skills -- visual perception (to recognize objects, scenes, obstacles), language understanding (to translate questions and instructions into actions), and navigation of potentially novel environments (to move and find things in a changing world). 
Much of the recent success in these areas is 
due to large neural networks trained on massive human-annotated datasets collected from the web.  However, this static paradigm of \emph{`internet vision'} is poorly suited for training embodied agents. By their nature, these agents engage in \emph{active perception} -- observing the world and then performing actions that in turn dynamically change what the agent perceives. What are needed then are richly annotated, photo-realistic environments where agents may learn about the consequence of their actions on future perceptions while performing high-level goals.

\begin{figure}[t]
    \centering
    \includegraphics[width=0.80\columnwidth]{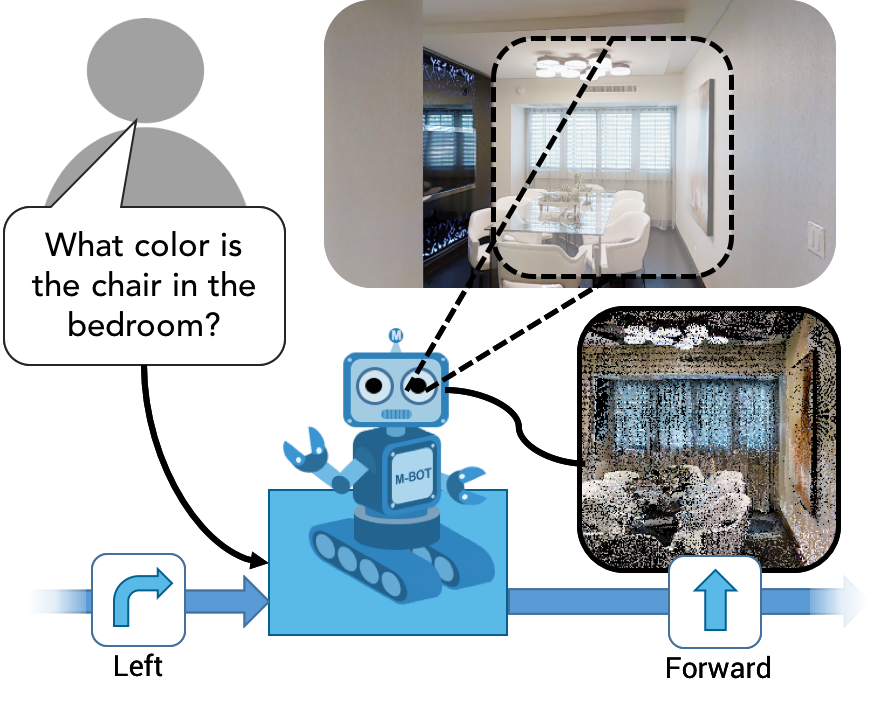}\vspace{-5pt}
    \caption{We extend \eqa \cite{embodiedqa} to \textit{photorealstic} environments, our agent is spawned in a \textit{perceptually} and \textit{semantically} novel environment and 
    tasked with answering a question about that environment.
    We examine the agent's ability to navigate the environment and answer the question by perceiving its environment through \pointclouds, RGB images, or a combination of the two.}
    \label{fig:teaser}
    \vspace{-0pt}
    \vspace{\captionReduceBot}
\end{figure}

To this end, a number of recent works have proposed goal-driven, perception-based tasks situated in simulated environments to develop such agents \cite{gupta_cvpr17, house3d,zhu2017target, zhu2017iccv,chaplot2017gated, mattersim, hermann2017grounded, embodiedqa, iqa2018, eqa_modular}. While these tasks are set in semantically realistic environments (\ie having realistic layouts and object occurrences), most are based in synthetic environments (on SUNCG~\cite{song_cvpr17} or Unity 3D models~\cite{juliani2018unity}) that are perceptually quite different from what agents embodied in the real world might experience. Firstly, these environments lack visual realism both in terms of the fidelity of textures, lighting, and object geometries but also with respect to the rich in-class variation of objects\footnote{To pervert Tolstoy, each ugly lamp is ugly in its own way.}. Secondly, these problems are typically approached with 2D perception (RGB frames) despite the widespread use of depth-sensing cameras (RGB-D) on actual robotic platforms \cite{huang2017visual, zeng2018robotic,AAMAS17-Zhang}. 

\xhdr{Contributions.} 
We address these points of disconnect by instantiating a large-scale, language-based navigation
task in photorealistic environments and by 
developing end-to-end trainable models with \pointcloud perception -- from raw 3D \pointclouds to goal-driven navigation policies.

Specifically, we generalize the recently proposed 
Embodied Question Answering (EmbodiedQA)~\cite{embodiedqa} task (originally proposed in synthetic SUNCG scenes~\cite{song_cvpr17}) to the photorealistic 3D reconstructions from Matterport 3D (MP3D)~\cite{Matterport3D}. 
In this task, an agent is spawned
at a random location in a novel environment (\eg a house) and asked to answer a question
(\myquote{What color is the car in the garage?}).
In order to succeed, the agent needs to navigate from egocentric vision alone (without an environment map), locate the entity in question (\myquote{car in the garage}),
and respond with the correct answer (\eg \myquote{orange}).

We introduce the \mpeqa dataset,
consisting of 1136 questions and answers grounded in 83  environments. 
Similar to \cite{embodiedqa}, our questions are generated from functional programs operating
on the annotations (objects, rooms, and their relationships) provided in MP3D; however, MP3D lacks color annotations for objects, which we collect from Amazon Mechanical Turk in order to generate \myquote{What color \ldots?} questions.
The MP3D environments provide significantly more challenging environments for our
agent to learn to navigate in due to the increased visual variation.

\newcommand\myeq{\mathrel{\overset{\makebox[0pt]{\mbox{\normalfont\tiny\sffamily def}}}{=}}}

We present a large-scale exhaustive evaluation of design decisions, training a total of 16 navigation models (2 architectures, 2 language variations, and 4 perception variations), 3 visual question answering models, and 2 perception models 
-- ablating the effects of perception, memory, and  goal-specification. Through this comprehensive analysis we demonstrate the complementary strengths of these perception modalities and highlight surprisingly strong baselines in the EmbodiedQA experimental setting. 

Our analysis reveals that the seemingly naive baselines, forward-only and random, are strong navigators in the default evaluation setting presented in \cite{embodiedqa} and challenging to beat, providing insight to others working in this space that models can perform surprisingly well without learning any meaningful behavior.
We also find that \pointclouds provide a richer signal than RGB images for learning obstacle avoidance, motivating continued study of utilizing depth information in embodied navigation tasks.

We find a novel weighting scheme we call \emph{Inflection Weighting} --  balancing the contributions to the cross-entropy loss between \textit{inflections}, where the ground truth action differs from the previous one, and non-inflections -- to be an effective technique when performing behavior cloning with a shortest path expert. 
We believe this technique will be broadly useful any time 
a recurrent model is trained %
on long sequences with an imbalance in symbol continuation versus symbol transition probabilities, \ie when $P(X_t = x \mid  X_{t-1} = x) >> P(X_t \neq x \mid  X_{t-1} = x)$.

To the best of our knowledge, this is the first work to explore end-to-end-trainable 3D perception
for goal-driven navigation in photo-realistic environments.
With the use of point clouds and realistic indoor scenes,
our work lays the groundwork for tighter connection between embodied vision and goal-driven navigation,
provides a testbed for benchmarking 3D perception models,
and hopefully
brings embodied agents trained on simulation one step closer to real robots
equipped with 2.5D \rgbd cameras.

%% file: sections/main/related_work.tex
\vspace{\sectionReduceTop}
\section{Related Work}
\vspace{\sectionReduceBot}

\xhdr{Embodied Agents and Environments.} 
End-to-end learning methods -- to predict actions 
directly from raw pixels~\cite{levine_jmlr16} -- 
have recently demonstrated strong performance. 
Gupta~\etal~\cite{gupta_cvpr17} learn to navigate via mapping and planning. Sadeghi~\etal~\cite{sadeghi2017cadrl} teach an agent to fly using simulated data and deploy it in the real world. Gandhi~\etal~\cite{gandhi2017} collect a dataset of drone crashes and train self-supervised agents to avoid obstacles.
A number of new challenging tasks have been proposed  including instruction-based  navigation~\cite{chaplot2017gated, mattersim}, target-driven navigation~\cite{zhu2017target,gupta_cvpr17}, embodied/interactive question answering~\cite{embodiedqa, iqa2018}, 
and task planning~\cite{zhu2017iccv}.

A prevailing problem in embodied perception is the lack of a standardized, large-scale, diverse, %
real-world benchmark -- essentially, there does not yet exist a COCO~\cite{mscoco} for embodied vision. 
A number of synthetic 3D environments have been introduced, such as DeepMind Lab~\cite{Beattie16_deepmind_lab} and VizDoom~\cite{vizdoom}. Recently, more visually stimulating and complex datasets have emerged which contain actionable replicas of 3D indoor scenes~\cite{house3d, savva2017minos, home2017, ai2thor}. These efforts make use of synthetic scenes~\cite{song2016ssc,ai2thor}, or scans of real indoor houses~\cite{Matterport3D,dai2017scannet} and are equipped with a variety of input modalities, \ie RGB, semantic annotations, depth, \etc.

The closest to our work is the EmbodiedQA work of  Das~\etal~\cite{embodiedqa}, who train 
agents to predict actions from egocentric RGB frames. 
While RGB datasets
are understandably popular for `internet vision', 
it is worth stepping back and asking
-- \emph{why must an embodied agent navigating in 3D environments be handicapped to perceive
with a single RGB camera?} 
We empirically show that point cloud representations are more effective for navigation in this task. Moreover, contrary to~\cite{embodiedqa, iqa2018} that use synthetic  environments, we extend the task to real environments sourced from~\cite{Matterport3D}.

\xhdr{3D Representations and Architectures.}
Deep learning has been slower to impact 3D computer vision than its 2D counterpart, in part due to the increased complexity of representing 3D data.
Initial success was seen with volumetric CNN's \cite{wu20153d, maturana2015voxnet, qi2016volumetric}.
These networks first discretize 3D space with a volumetric representation and
then apply 3D variants of operations commonly found in 2D CNN's -- convolutions, pooling, etc.
Volumetric representations are greatly limited due to the sparsity of 3D data and the computational cost of
3D convolutions.  
Recent works on 3D deep learning have proposed architectures that operate directly on \pointclouds. Point clouds are a challenging input for deep learning as they are naturally a set of points with no canonical ordering.
To overcome the ordering issue, some utilize symmetric functions, PointNet(++) \cite{qi2017pointnet,qi2017pointnet++}, and A-SCN \cite{xie2018attentional}.  Others have used clever internal representations, such as 
SplatNet \cite{su2018splatnet},  Kd-Net \cite{klokov2017escape}, and O-CNN \cite{wang2017ocnn}.

%% file: sections/main/data.tex
\vspace{\sectionReduceTop}
\section{Questions in Environments}
\vspace{\sectionReduceBot}

\begin{figure}
\setlength{\fboxsep}{0pt}
\setlength{\fboxrule}{1pt}
\begin{subfigure}{0.22\textwidth}
\centering
\fbox{\includegraphics[width=0.95\textwidth]{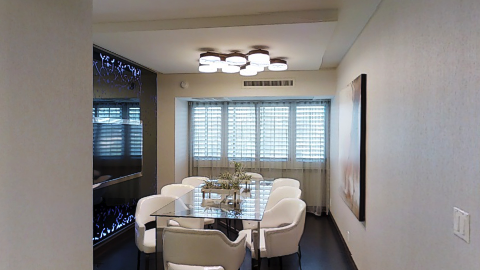}}\\[-4pt]
\caption{RGB Panorama}
\label{fig:pano}
\end{subfigure}
\hfill
\begin{subfigure}{0.22\textwidth}
\centering
\fbox{\includegraphics[width=0.95\textwidth]{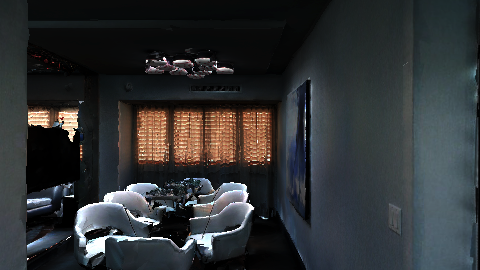}}\\[-4pt]
\caption{Mesh Reconstruction}
\label{fig:mesh}
\end{subfigure}
\hfill
\begin{subfigure}{0.22\textwidth}
\centering
\fbox{\includegraphics[width=0.95\textwidth]{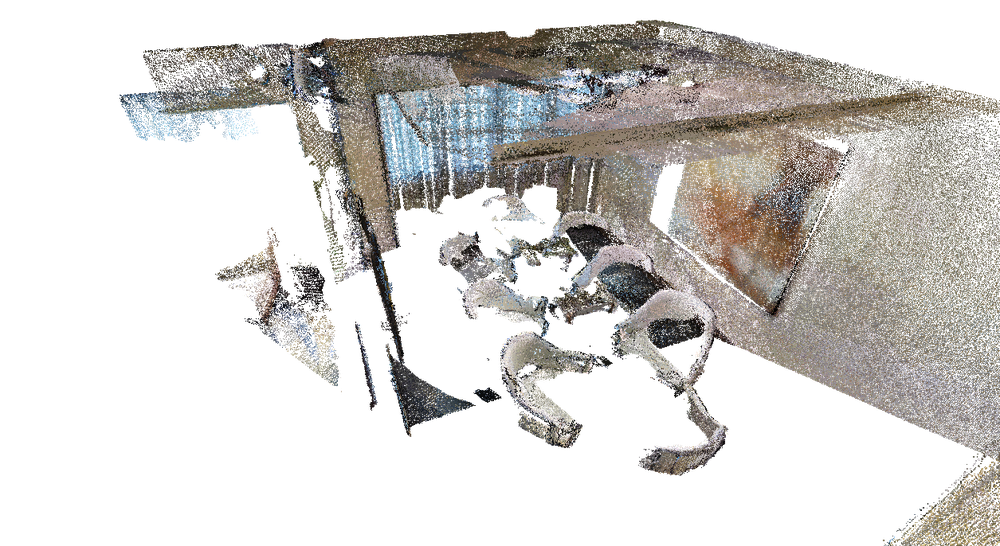}}\\[-4pt]
\caption{Point Cloud}
\label{fig:pc}
\end{subfigure}
\hfill
\begin{subfigure}{0.22\textwidth}
  \centering
\fbox{\includegraphics[width=0.95\textwidth]{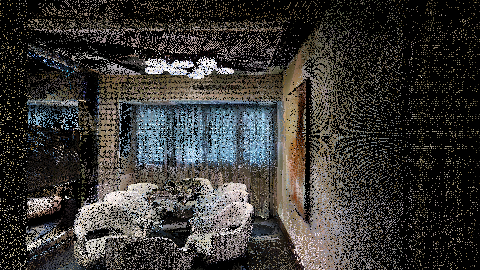}}\\[-4pt]
\caption{RGB-D Render}
\label{fig:render}
\end{subfigure}\\[-8pt]
\caption{Illustration of mesh construction errors and what point clouds are able to correct.  Notice the warping of flat surfaces, the extreme differences in color, and texture artifacts from reflections.}
\vspace{\captionReduceBot}
\end{figure}

In this work, we instantiate the Embodied Question Answering (EQA) \cite{embodiedqa} task
in realistic environments from the Matterport3D dataset \cite{Matterport3D}.

\vspace{\subsectionReduceTop}
\subsection{Environments}
\vspace{\subsectionReduceBot}

The Matterport3D dataset consists of 90 home environments captured
through a series of panoramic RGB-D images taken by a Matterport
Pro Camera (see sample panoramas in \figref{fig:pano}).
The resulting point clouds are aligned and used to reconstruct
a 3D mesh (like those shown in \figref{fig:mesh}) that is then annotated with semantic labels. The Matterport3D dataset is densely annotated with semantic segmentations
of 40 object categories for $\sim$50,000 instances.
Room type is annotated for over 2050 individual rooms.

These reconstructions offer high degrees of perceptual realism but are not perfect however and sometimes suffer from discoloration and unusual geometries such as holes in surfaces. 
In this work, we examine both RGB and RGB-D perception in these environments. For RGB, we take renders from the mesh reconstructions and for \pointclouds we operate directly on the aligned point clouds. 
\figref{fig:pc} and \figref{fig:render} show the point cloud rendered
 for an agent looking at the scene shown in \figref{fig:pano}.

\xhdr{Simulator.} To enable agents to navigate in MatterPort3D
environments, we develop a simulator based on MINOS \cite{savva2017minos}.
Among other things, MINOS provides occupancy checking, RGB frame rendering from the mesh,
and shortest path calculation (though we reimplement this for higher
accuracy and speed). It does not however provide access to the underlying point clouds. In order to render 2.5D \rgbd frames, we
first construct a global point cloud from all of the panoramas
provided in an environment from the dataset.  Then, the agent's current position, camera parameters (field of view, and
aspect ratio), and the mesh reconstruction are used to determine which points are within its view. 
See the supplementary for full details on this.

\vspace{\subsectionReduceTop}
\subsection{Questions}
\vspace{\subsectionReduceBot}

Following \cite{embodiedqa}, we programmatically generate templated questions
based on the Matterport3D annotations, generating questions of the following three types:

\vspace{5pt}
\begin{compactitem}[\hspace{3pt}]
\item \textbf{location:} \emph{What room is the} \texttt{<OBJ>} \emph{located in?}
\item \textbf{color:} \emph{What color is the} \texttt{<OBJ>} \emph{?}
\item \textbf{color\_room:} \emph{What color is the} \texttt{<OBJ>} \emph{in the} \texttt{<ROOM>} \emph{?}
\end{compactitem}
\vspace{5pt}

\noindent 
While EQA \cite{embodiedqa} included a forth question type \textbf{prepositions}, we found those 
questions in MP3D to be relatively few, 
with strong biases 
in their answer, thus we do not include them. 

While room and object annotations and positions supporting the three question types above are available in MP3D, 
human names for object colors are not. 
To rectify this, we collect the dominant color of each  object from workers on 
Amazon Mechanical Turk (AMT).
Workers are asked to
select one of 24 colors for each object.  The color palette was created by starting with Kenneth
Kelly's 22 colors of maximum contrast \cite{kelly1965twenty} and adding 2 additional colors (off-white and slate-grey) due to their prevalence in indoor scenes. Overall, the most reported color was gray.
For each object, we collect 5 annotations and take the majority vote, breaking ties based on object color priors.
 We include details of the AMT interface in the supplementary.

Following the protocol in \cite{embodiedqa}, we filter out questions that have a low entropy in distribution over 
answers across environments \ie 
peaky answer priors --  \eg the answer to \myquote{What room is the shower in?} is nearly always \myquote{bathroom} -- 
to ensure that questions in our dataset require the agent to navigate and perceive to answer accurately. We remove
rooms or objects that are ambiguous (\eg ``misc'' rooms) or structural
(\eg ``wall'' objects). %
Below are the objects and rooms that
appear in our generated questions:\\[-8pt]

{\small 
\centering
\begin{minipage}[l]{0.95\columnwidth}
\textbf{Objects:} \emph{shelving, picture, sink, clothes,
appliances, door, plant,
furniture, fireplace, chest of drawers,
seating, sofa, table,
curtain, shower, towel,
cushion, blinds, counter,
stool, bed, chair,
bathtub, toilet, cabinet}
\end{minipage}

\begin{minipage}[l]{0.95\columnwidth}

\textbf{Rooms:} \emph{family room, closet, spa,
dinning room, lounge, gym,
living room, office, laundry room,
bedroom, foyer, bathroom,
kitchen, garage, rec room,
meeting room, hallway, tv room}\\[-7pt]
\end{minipage}

}

\noindent In total, we generate $\sim$1100 questions across 83
home environments (7 environments resulted in no questions after filtering).
Note that this amounts to $\sim$13 question-answer pairs per environment compared to $\sim$12 per scene in~\cite{embodiedqa}. Color\_room questions make up the
majority of questions. These questions require searching the environment
to find the specified object in the specified room.  Whereas \cite{embodiedqa} requires both the object and the room to be unique within the environment, we only require the (object, room) pair to be unique, thereby giving the navigator significantly less information about the location of the object.

We use the same train/val/test split of environments as in MINOS
\cite{savva2017minos}. 
Note that in \cite{embodiedqa}, the test environments 
differ from train only in the layout of the objects; 
the objects themselves have been seen during training. 
In MP3D-EQA, the agents are tested on entirely new 
homes, thus may come across entirely new objects --  testing %
semantic \emph{and} perceptual generalization. 
\reftab{tab:splits} shows the distribution
of homes, floors, and questions across these splits. We restrict agent
start locations to lie on the same floor as question targets
and limit episodes to single floors.

\begin{table}
\centering
\resizebox{0.9\columnwidth}{!}{
\begin{tabular}[t]{c c c c c}
\toprule
    & Homes & Floors & Total Qns. & Unique Qns. \\
\midrule
train	& 57			& 102		& 767        	& 174\\
val     & 10			& 16			& 130			& 88\\
test	& 16			& 28			& 239			& 112\\
\bottomrule
\end{tabular}}
\vspace{-5pt}
\caption{Statistics of splits for EQA in Matterport3D}
\label{tab:splits}
\vspace{\captionReduceBot}
\end{table}

	%

%% file: sections/main/approach.tex
\vspace{\sectionReduceTop}
\section{Perception for Embodied Agents}
\vspace{\sectionReduceBot}

Agents for EmbodiedQA must understand the given question, perceive
and navigate their surroundings collecting information, and answer
correctly in order to succeed.
Consider an EmbodiedQA agent that navigates by predicting an
action $a_t$ at each time step $t$ based on its
trajectory of past observations and actions $\sigma_{t-1} =
(s_1, a_1, s_2, a_2, \dots, s_{t-1}, a_{t-1})$, the current state $s_t$, and the question $Q$.
There are many important design decisions for such a model -- action
selection policy, question representation, trajectory encoding, and
observation representation. In this work, we focus on the observation
representation -- \ie perception -- in isolation and follow the architectural pattern in \cite{embodiedqa} for the remaining components. In this section,
we describe our approach and recap existing model details.

\vspace{\subsectionReduceTop}
\subsection{Learning Point Cloud Representations}
\vspace{\subsectionReduceBot}

\label{sec:percept}

\begin{figure*}
    \centering
    \includegraphics[width=0.9\textwidth]{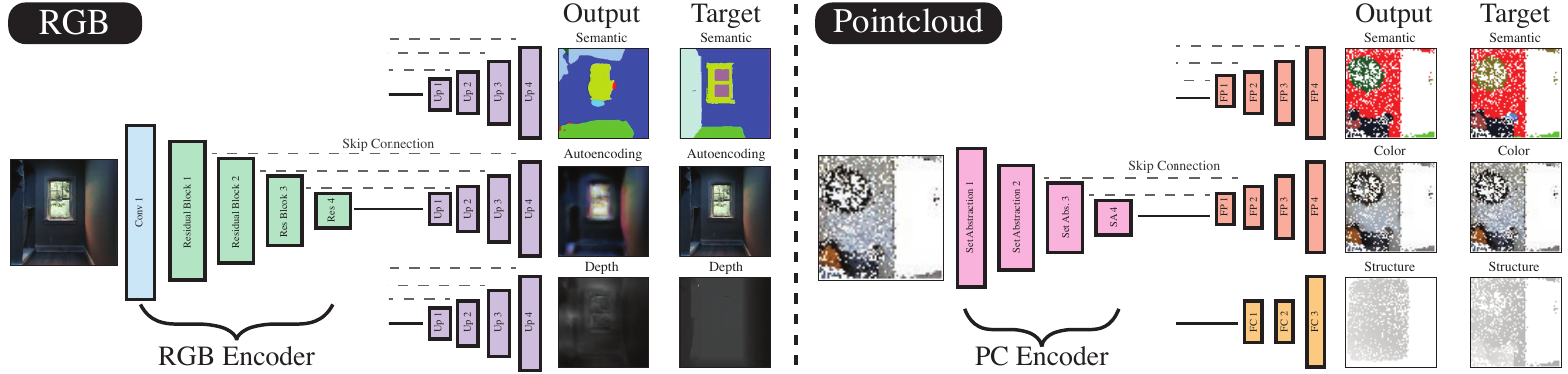}
    \caption{The visual encoders a trained using three pertaining tasks to imbue their scene representations with information about semantics (segmentation), color (autoencoding), and structure (depth).  All decoder heads share the same encoder.  Up-sampling for RGB (Up \#) is done with bi-linear interpolation. Upsampling for pointclouds (FP \#), is achieved with Feature Propagation layers \cite{qi2017pointnet++}.  After pretraining, the decoders are discarded, and the encoder is treated as a static feature extractor.}
    \label{fig:pretrain}
    \vspace{\captionReduceBot}
\end{figure*}

Consider a \pointcloud $P \in \mathcal{P}$ which is an unordered set of
points in 3D space with associated colors, \ie
$P=\{(x_m,y_m,z_m,R_m,G_m,B_m)\}_{m=1}^M$. To enable a neural
agent to perceive the world using \pointclouds, we must learn a
function $f: \mathcal{P} \rightarrow \mathbb{R}^{d}$ that maps
a \pointcloud to an observation representation. To do this, we leverage a widely used 3D architecture, PointNet++ \cite{qi2017pointnet++}. 

\xhdr{PointNet++.} At a high-level, PointNet++ alternates
between spatial clustering and feature summarization -- resulting in
a hierarchy of increasingly coarse point clusters with associated
feature representations summarizing their members. This approach
draws an analogy to convolution and pooling layers in standard
convolutional neural networks. %

More concretely, let $\{p_1^i, ..., p^i_{N_i}\}$ be the set of $N_i$ points
at the $i^\text{th}$ level of a PointNet++ architecture and $\{h_1^i, ..., h^i_{N_i}\}$
be their associated feature representations (\eg RGB values for the
input level). To construct the $i+1^\text{th}$ level, $N_{i+1}$ centroids
$\{p_1^{i+1}, ..., p_{N_{i+1}}^{i+1}\}$ are sampled from level $i$
via iterative farthest point sampling (FPS) -- ensuring even
representation of the previous layer. These centroids will
make up the points in level $i+1$ and represent their local areas.
For each centroid $p_k^{i+1}$, the $K$ closest points within a
max radius are found and a symmetric learnable neural
architecture \cite{qi2017pointnet},  composed of a series of
per-point operations (essentially 1-by-1 convolutions) and
a terminal max-pool operation, is applied to this set of associated points to produce the summary representation $h_k^{i+1}$.
These clustering and summarization steps (referred to as Set Abstractions in \cite{qi2017pointnet++}) can be repeated
arbitrarily many times. In this work we use a 3 level
architecture with $N^1=1045$, $N^2=256$, and $N^3=64$. We
compute a final feature with a set of 1-by-1 convolutions and a max-pool over the 3rd level point features and denote this network as $f(\cdot)$.

Given an input point cloud $P_t$ from an agent's view at time $t$,
we produce a representation $s_t = f(P_t)$ where $s_t \in
\mathbb{R}^{1024}$.  However, point clouds
have an interesting property -- as an agent navigates an environment the
number of points it perceives can vary. This may be due to sensor limitations
(\eg being too close or too far from objects) or properties of the observed
surfaces (\eg specularity). While the encoder $f$ is invariant to
 the number of input points, representations drawn from few
 supporting points are not likely to be good representations of a scene.
 For a navigation or question-answering agent, this
 means there is no way to discern between confident and unconfident
 or erroneous observation.
To address this, we divide the range spanning the possible number of points in any given point cloud -- $[0, 2^{14}]$ --
into 5 equal sized bins and represent these bins as 32-d feature vectors that encode the sparsity of a point cloud.
Now, given a point cloud $P_t$ with $|P_t|$ points, we retrieve its corresponding sparsity
embedding $c_{t}$ and produce a final encoding $[s_t, c_{t}] \in \mathbb{R}^{1056}$ that is used by the agent
for navigation and question-answering.

\xhdr{Visual Pretraining Tasks.} To train the encoder architecture to
extract semantically and spatially meaningful representations of agent views, we
 introduce three pretraining tasks based on the annotations provided in
 Matterport3D. Specifically, these tasks are:
\begin{compactenum}[\hspace{2pt}--]

\item \textbf{Semantic Segmentation} in which the model predicts the object annotation for each
  point, $y^s_i$, from the summarized representation $s_i = f(P)$.
  We train a PointNet++ feature propagation network $g_\text{s}(\cdot)$
  to minimize the cross-entropy between $y^s_i$ and $\hat{y}^s_i = g_\text{s}(f(P))$ \cite{qi2017pointnet++}.
  This encourages the encoder, $f(\cdot)$, to include information about which objects are in the
  frame.

\item \textbf{Color Autoencoding} mirrors the semantic segmentation task.  However, the
  network $g_\text{c}(\cdot)$ is now trained to minimize the smooth-L1 loss between
  $y^c_i$ and $\hat{y}^c_i = g_\text{c}(f(P))$. This task encourages
  the encoder $f(\cdot)$ to capture holistic information
  about the colors in the scene.

\item \textbf{Structure Autoencoding} where point coordinates must be recovered from the summarized representation, \ie
 	$\{(x_i,y_i,z_i, R_i, G_i, B_i)\}_{i=1}^N \rightarrow \{(x_i,y_i,z_i)\}_{i=1}^N$.
 	We implement this decoder as a multi-layer perceptron that
 	regresses to the $N\times3$ spatial coordinates.
 	As in \cite{achlioptas_arxiv2017}, we use the
	earth-movers distance as the loss function.

\end{compactenum}
We demonstrate these tasks in  \reffig{fig:pretrain}.
These tasks encourage the model features to
represent colors, objects, and
spatial information including free-space and depth
that are essential to navigation.
We collect $\sim$100,000 frames from
Matterport3D using our simulator and train
the \pointcloud encoder for these tasks. 
We discard the decoder networks after training, and use the encoder $f$ as a fixed feature extractor.

\xhdr{RGB Image representations.} We utilize ResNet50~\cite{he2016deep} trained using an analogous set of tasks (semantic segmentation, autoencoding, and depth prediction) to learn a representation for egocentric $224\times224$ RGB images as in \cite{embodiedqa}.  \newcontent{We find that ResNet50 is better able to handle the increased visual complexity of the \matterport environments than the shallow CNN model used in Das \etal.}
We provide further details about perception model and decoder architectures in the supplement.

\vspace{\subsectionReduceTop}
\subsection{Navigation and Question Answering}
\vspace{\subsectionReduceBot}

We now provide an overview of the navigation and
question answering models we use in this work.

\xhdr{Question Encoding.} In order to succeed at navigation and question answering, it is important for an
embodied agent to understand the queries it is being tasked with answering. We use two layer LSTMs with 128-d hidden states to encode questions.
The question encoding for navigation and question answering are learned separately.

\xhdr{Question Answering Models.} We experimented with three classes of question answering models:

\begin{compactitem}[\hspace{2pt}--]
\item \textbf{Question-only} We examine the question-only baselines proposed in \cite{embodiedqa} --  a small classification network that predicts the answer using just the question encoding.  We also examine the recently proposed question-only baselines in \cite{anand2018blindfold}
-- a simple nearest neighbors approach and a bag-of-words with a softmax classifier.
\item \textbf{Attention}  This is the best performing VQA model from \cite{embodiedqa}. It computes question-guided attention over the features of the last five frames observed by the agent before stopping, followed by element-wise product between the attended feature and question encoding to answer; and
\item \newcontent{\textbf{Spatial Attention} utilizes the bag-of-words encoder proposed in \cite{anand2018blindfold} to compute spatial attention over the last-frame.  We use scaled dot-product attention \cite{vaswani2017attention} over the feature map, perform an element-wise product between attended features and the question feature, and predict an answer. This model only uses RGB inputs.}
\end{compactitem}

\xhdr{Navigation Models.} 
We consider two baseline navigators: 
\begin{compactitem}[\hspace{2pt}--]
\item \textbf{Forward-only (Fwd)} which always predicts forward.
\item \textbf{Random} which uniformly chooses one of \texttt{forward}, \texttt{turn-left}, and \texttt{turn-right} at every time step.
\end{compactitem}

\noindent
We consider two navigation architectures:
\begin{compactitem}[\hspace{2pt}--]
\item \textbf{Reactive (R)} which is a simple feed-forward network that takes a concatenation of the embedding of the five most
recent visual observations as input to predict an action. As such, this is a memory-less navigator.%
\item \textbf{Memory (M)} which is a two-layer GRU-RNN that takes the encoding(s) of the current observation and previous action as inputs to predict the current action.
\end{compactitem}

\noindent For each navigation architecture, we examine the combination of our 4 different perception variations, \texttt{None} (\ie a blind model as suggested by Thomason \etal \cite{thomason2018shifting}), \texttt{PC}, \texttt{RGB}, and \texttt{PC+RGB}, with the 2 different language variations, \texttt{None} and \texttt{Question}. 
For reactive models that utilize the question, we incorporate the question embedding by concatenation with the visual embedding.  For memory models, the question embedding is an additional input to the GRU. Due to the highly correlated observations our agents see during training, we utilize Group Normalization layers~\cite{wu2018group}~
in our navigation models.
The action space for all our navigation models is \texttt{forward}, \texttt{turn-left}, \texttt{turn-right}, and \texttt{stop}.

\vspace{\subsectionReduceTop}
\subsection{Imitation Learning from Expert Trajectories}
\label{sec:expert}
\vspace{\subsectionReduceBot}

To train our models, we first create a static dataset of agent trajectories
by generating training episodes based on shortest-paths from agent spawn
locations to the best view of the object of interest. 
For example, if a question asks
\myquote{What color is the sofa in the living room?}, we spawn an agent
randomly in the environment in the same floor as the target object -- the sofa -- and compute the shortest navigable path to the best view of the sofa.  The best view of the sofa is determined by exhausting all possible view positions within a reasonable radius of the target object.  The quality of a view is determined by the intersection over union of a pre-determined bounding box and the segmentation mask of the target. In normalized image coordinates, the bounding box's top left corner is at (0.25, 0.25) and it has a height of 0.6 and a width of 0.5. We use this metric instead of simply maximizing the number of visible pixels in the segmentation mask to maintain context of the object's surroundings.

To provide enough data to overcome the complexity of Matterport3D environments,
we generate $\sim$11,796 such paths in total
(corresponding to approximately $\sim$15 episodes per question-environment pair,
each for a different spawn location of the agent).
For computational efficiency in the large Matterport3D environments, we compute
shortest paths in continuous space using LazyTheta* \cite{nash2010lazy} and greedily generate agent actions to follow that path,
rather than directly searching in the agent's action space.

\xhdr{Perception.} We use the frozen pre-trained perception models as described in \secref{sec:percept}. For \texttt{PC+RGB} models we simply concatenate both visual features.

\xhdr{Question Answering.}  The question answering models are trained to predict the ground truth answer from a list of $53$ answers using Cross Entropy loss. 
The models with vision use the ground-truth navigator during training.

\vspace{\subsectionReduceTop}
\subsection{Imitating Long Trajectories Effectively}
\vspace{\subsectionReduceBot}

All navigation models are trained with behavior cloning where they are made to mimic the ground truth, shortest path agent trajectories. That is to say the agents are walked through the ground truth trajectory observing the corresponding frames (though reactive models retain only the last five) up until a given time step and then make an action prediction. Regardless of the decision, the agent will be stepped along the ground truth trajectory and repeat this process. One challenge with this approach is that relatively unintelligent policies can achieve promising validation loss without really learning anything useful -- one such strategy simply repeats the previous ground truth action. Insidiously, these models achieve very high validation accuracy for action prediction but miss every transition between actions!

\xhdr{Inflection Weighting.} To combat this problem and encourage agents to focus on important decisions along the trajectory, we introduce a novel weighting scheme we call Inflection Weighting. Conceptually, we weight predictions at time steps more heavily if the ground truth action differs from the previous one -- that is if the time step is an inflection point in the trajectory. More formally, we define a per-time step weight 
\vspace{-5pt}
\begin{align}
w_t = \begin{cases} 
      \frac{N}{n_{I}} & a_{t-1} \neq a_{t}  \\
      1 & \text{otherwise}
   \end{cases}
\end{align}
\noindent where $N/n_I$ is the inverse frequency of inflection points (approximately 5.7 in our dataset). We can then write an inflection weighted loss between a sequence of predictions $\hat{Y}$ and a ground truth trajectory $A$ over as:
\vspace{-5pt}
\begin{align}
\ell_{IW}(\hat{Y},A) = \frac{1}{\sum_{t=1}^T w_t}~ \sum_{t=1}^T w_t\ell\left(\hat{y}_t, a_t\right)
\end{align}
\noindent where $\ell(\cdot, \cdot)$ is the task loss -- cross-entropy in our setting. We define the first action, $t=1$, to be an inflection. In practice, we find inflection weighting leads to significant gains in performance for recurrent models.

\newcontent{Inflection weighting may be viewed as a generalization of the class-balanced loss methods that are commonly used in supervised 
learning under heavily imbalanced class distributions (\eg in semantic segmentation~\cite{eigen2015predicting}) 
for a particular definition of a `class' (inflection or not).}

%% file: sections/main/results.tex
\vspace{\sectionReduceTop}
\section{Experiments and Analysis}
\vspace{\sectionReduceBot}

\newcommand{\dt}{$\mathbf{d_T}$\xspace}
\newcommand{\iout}{$\mathbf{IoU_T}$\xspace}
\newcommand{\qa}{$\mathbf{QA}$\xspace}
\newcommand{\dmin}{$\mathbf{d_{min}}$\xspace}
\newcommand{\pcollide}{$\mathbf{\%_{collision}}$\xspace}

We closely follow the experimental protocol of  Das~\etal~\cite{embodiedqa}. 
All results here are reported on novel test environments.
Agents are evaluated on their performance 10, 30, or 50
primitive actions away from the question target, corresponding to distances of 
0.35, 1.89, and 3.54 meters respectively.
One subtle but important point is that 
to achieve these distances the agent is first randomly spawned within the environment, and then the agent is \emph{walked along the shortest path to the target} until it is the desired distance from the target (10, 30, or 50 steps).

We perform an exhaustive evaluation of design decisions, training a total of 16 navigation models (2 architectures, 2 language variations, and 4 perception variations), 3 visual question answering models, and 2 perception models.  %

\vspace{\subsectionReduceTop}
\subsection{Metrics}
\vspace{\subsectionReduceBot}

\xhdr{Question Answering.} For measuring question answering performance, we report the top-1 accuracy, \ie
did the agent's predicted answer match the ground truth or not.

\xhdr{Navigation.} For navigation, we report the distance to the target object from where the agent is
spawned ($\mathbf{d_0}$) for reference, measure distance to the target object
upon navigation completion \dt (lower is better), and the percentage of actions that result in a collision with an obstacle \pcollide (lower is better). 
All the distances are geodesic, \ie measured along the shortest path.

We propose a new metric, \iout (higher is better), to evaluate
the quality of the view of the target the agent obtains at the end of navigation.  %
We compute the intersection-over-union (IoU) score between the ground-truth target segmentation and the same centered bounding box used to select views during dataset generation (see \secref{sec:expert}). To compensate for object size, we divide by the best attainable IoU for the target object. We define \iout as the maximum of the last $N$ IoU scores.  We set $N$ to $5$ as the VQA model receives the last $5$ frames.

\vspace{\subsectionReduceTop}
\subsection{Results and Analysis}
\vspace{\subsectionReduceBot}

\newlength{\plotwidth}
\setlength{\plotwidth}{0.975\columnwidth}

\begin{figure}
    \centering
    \includegraphics[width=\plotwidth]{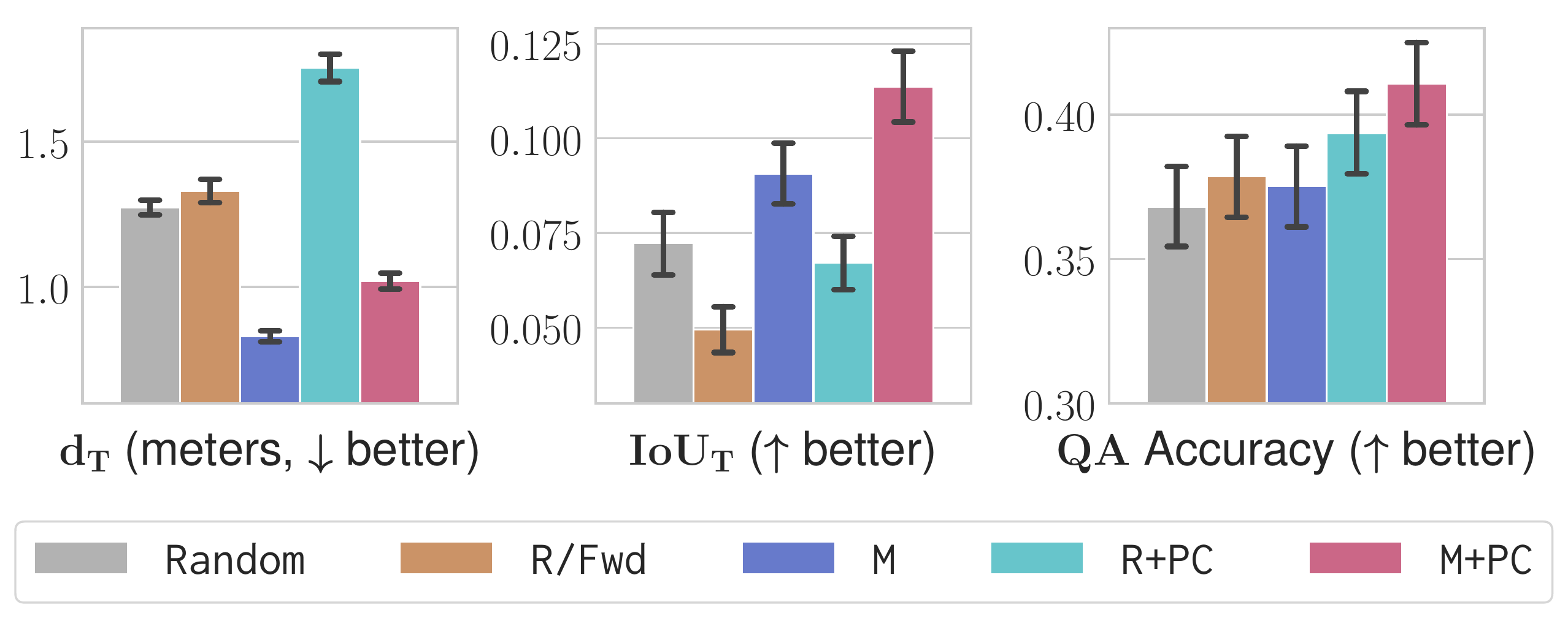}
    \caption{Models with memory significantly outperform their memory-less counterparts. Surprisingly, the baselines, random and forward-only, and a vision-less navigator with memory perform very well.}
    \label{fig:memory_helps}
    \vspace{\captionReduceBot}
\end{figure}

\begin{figure}
    \centering
    \includegraphics[width=\plotwidth]{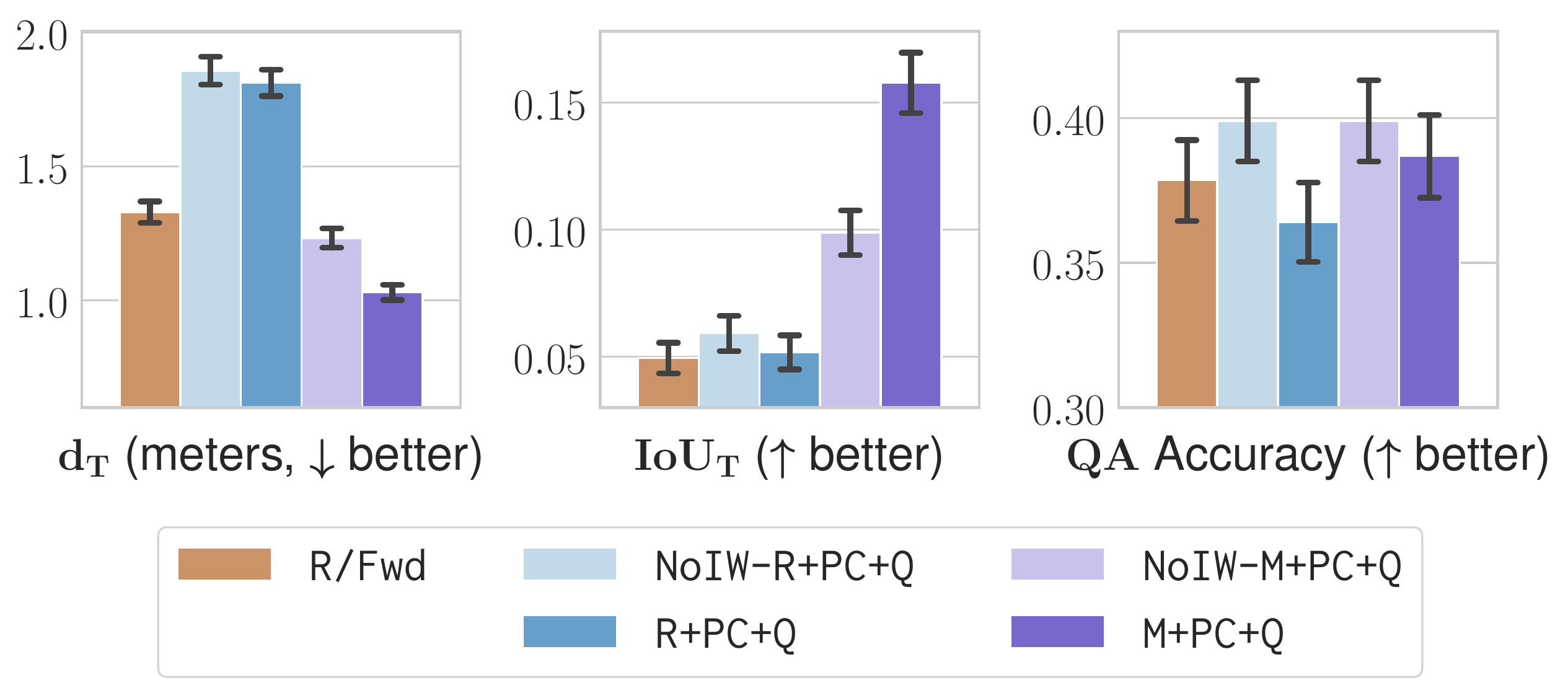}
    \caption{Models trained with inflection-weighted cross-entropy loss 
    significantly outperform
    their unweighted cross-entropy counterparts
    and the baselines.}
    \label{fig:iw_plot}
    \vspace{\captionReduceBot}
\end{figure}

\begin{figure*}
    \centering
    \includegraphics[width=\textwidth]{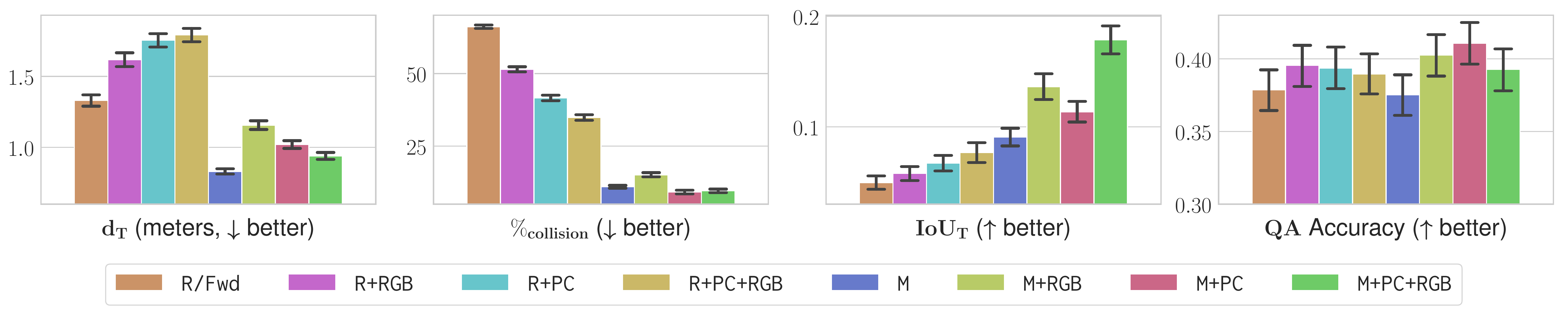}
    \caption{Vision generally hurts distance-based navigation metrics.  However metrics that are dependent on the navigators ability to look in a particular direction (\iout and \qa) generally improve, and the models collide with the environment less.}
    \label{fig:vision}
    \vspace{\captionReduceBot}
\end{figure*}

\begin{figure}
    \centering
    \includegraphics[width=0.95\plotwidth]{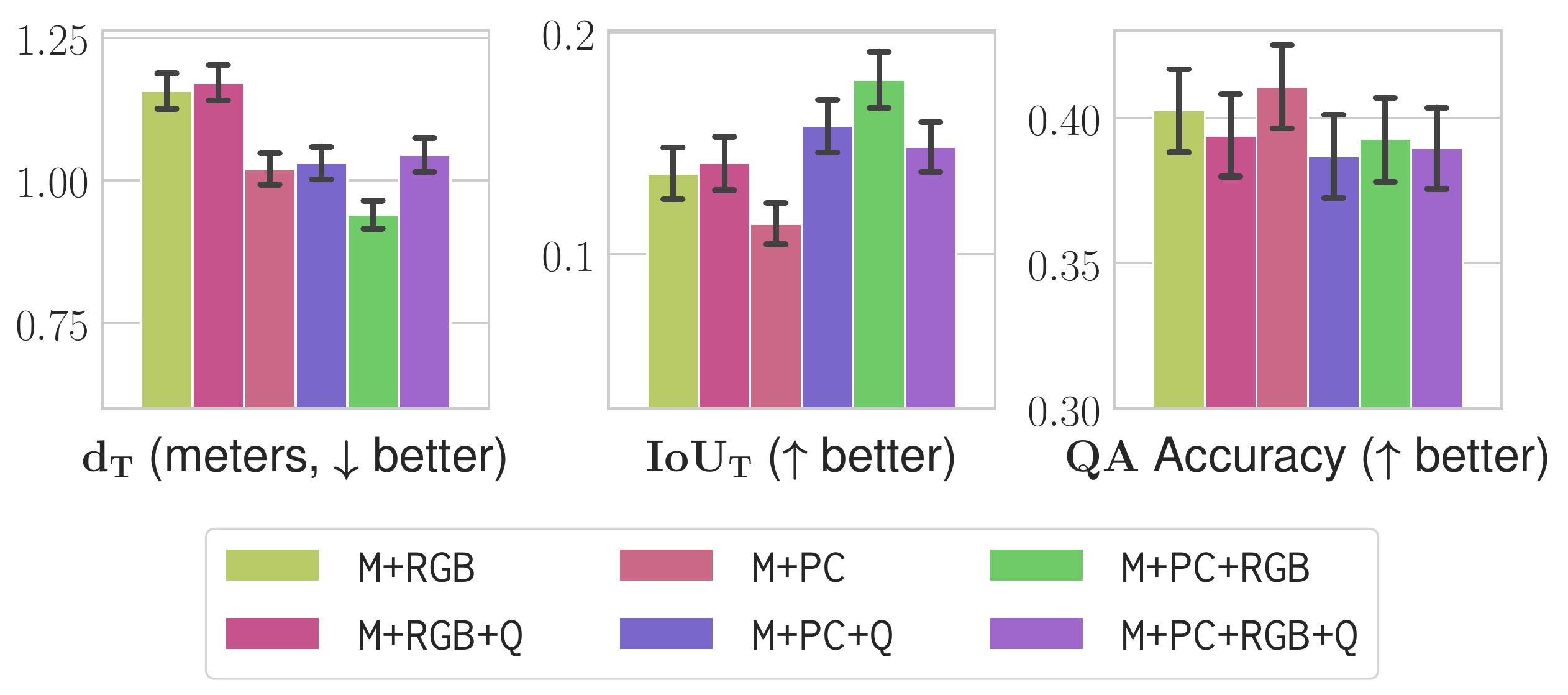}
    \caption{Comparison of memory navigation models with and without the question.  Interestingly, adding the question doesn't appear to aid models trained with behavior cloning.}
    \label{fig:question_comp}
    \vspace{\captionReduceBot}
\end{figure}

\noindent\textbf{Question Answering.} 
The top-1 accuracy for different answering modules on the validation set using the ground-truth navigator is shown below.\\[-13pt]

\begin{center}
\resizebox{0.6\columnwidth}{!}{
\begin{tabular}{@{}l c@{}}
 \toprule
 & Top-1 (\%) \\ 
 \midrule
\newcontent{\texttt{spatial+RGB+Q}} & \textbf{46.2}\\
\texttt{attention+RGB+Q} & 40.0\\
\texttt{attention+PC+RGB+Q} & 38.4 \\
\texttt{attention+PC+Q} & 36.1 \\
\texttt{lstm-question-only} & 32.8 \\
\texttt{nn-question-only} \cite{anand2018blindfold} & 35.4 \\
\texttt{bow-question-only} \cite{anand2018blindfold} & 38.3 \\
\bottomrule
\end{tabular}}
\end{center}
\noindent In-order to compare \qa performance between navigators, we report all \qa results with the best-performing module -- \texttt{spatial+RGB+Q} -- regardless of the navigator.

\xhdr{Navigation.} 
We use the following notation to specify our models:  For the base architecture, \texttt{R} denotes reactive models and \texttt{M} denotes memory models.  The base architectures are then augmented with their input types, \texttt{+PC}, \texttt{+RGB}, and \texttt{+Q}. So a memory model that utilizes \pointclouds (but no question) is denoted as \texttt{M+PC}. Unless otherwise specified (by the prefix \texttt{NoIW}), models are trained with inflection weighting.
We denote the two baseline navigators, forward-only and random, as \texttt{Fwd} and \texttt{Random}, respectively.

Due to the large volume of results, we present key findings and analysis here (with $T_{-30}$) and, for the intrepid reader, \supp{provide the full table (with 300+ numbers!) in the supplement.} 
We make the following observations:

\xhdr{Forward-only is a strong baseline.}  One of the side-effects of the evaluation procedure proposed in \cite{embodiedqa} is that the agent is commonly facing the correct direction when it is handed control. This means the right thing to do to make progress is to go forward. As a result, a forward-only navigator does quite well, see \reffig{fig:memory_helps}. Forward-only also tends to not overshoot too much due to its `functional stop': continually running into an obstacle until the max step limit is reached. Our vision-less reactive models 
(\texttt{R/Fwd} and \texttt{R+Q/Fwd}) learn to only predict forward, the most frequent action. %

\reffig{fig:memory_helps} also shows that the random baseline is a deceptively strong baseline. 
The lack of a \texttt{backward} action, 
and \texttt{left} and \texttt{right} cancelling each other out in expectation, 
results in random essentially becoming forward-only.

\xhdr{Inflection weighting improves navigation.}  We find inflection
weighting to be crucial for training navigation models with behavior 
cloning of a shortest-path expert; see~\reffig{fig:iw_plot}. While we see some 
improvements with inflection weighting for most models, memory models 
reap the greatest benefits -- improving significantly on both \dt
and \iout. Interestingly, these gains do not translate into improved 
\qa accuracy. While we have only utilized this loss for behavior cloning, 
we suspect the improvements seen from inflection weighting will transfer to models that are fine-tuned with reinforcement
learning as they begin with better performance.

\xhdr{Memory helps.} \reffig{fig:memory_helps} shows that models with memory are better navigators than their reactive counter parts.  Surprisingly, a vision-less navigator with memory performs very well at distance based navigation metrics.  Like a vision-less reactive navigator (forward-only), a vision-less memory navigator is only able to learn priors on how shortest paths in the dataset tend to look, however memory allows the model to count and therefore it is able to stop and turn. %
    
\xhdr{Vision helps gaze direction metrics.} \reffig{fig:vision} shows the effect of adding vision to both reactive and memory models.  The addition of vision leads to improvements on \iout and \qa, however, the improvements in \iout do not translate directly improvement on \qa.  This is likely due to naive VQA models.  Models with vision also tend to collide with the environment less often, as can be seen by \pcollide usually being lower.  

\xhdr{Vision hurts distance metrics.} Surprisingly, adding vision hurts distance based navigation metrics (\dt). For reactive models, adding vision causes the models to collide significantly less frequently, resulting in a loss of the `functional stop' that forward-only uses, \ie continually colliding until the step limit is reached.  For memory models, the story isn't as clear; however, memory models with vision stop less often and thus have a higher average episode length than their vision-less counterpart, which causes them to overshoot more often.  We suspect this is because they learn a more complex function for stopping than the simple counting method used by vision-less memory models and this function is less able to handle errors during navigation.
    
\xhdr{Question somewhat helps.} \reffig{fig:question_comp} provides a comparison of \texttt{M}+\texttt{PC} and \texttt{M}+\texttt{RGB} and \texttt{M}+\texttt{PC}+\texttt{RGB} with and without the question (\texttt{Q}).  Interestingly, we do not see large improvements when providing models with the question.
Given how much \textbf{color\_room} dominates our dataset, it seems reasonable to expect that telling the navigation models which room to go to would be a large benefit.  We suspect that our models are not able to properly utilize this information due to limitations of behavior cloning.  Models trained with behavior cloning never see mistakes or exploration and therefore never learn to correct mistakes or explore.  

 \xhdr{\texttt{PC+RGB} provides the best of both worlds.} \reffig{fig:vision} also provides a comparison of the three different vision modalities.  The general tend is that \pointclouds provided a richer signal for obstacle avoidance (corresponding to lower \pcollide values), while RGB provides richer semantic information (corresponding to a higher \iout and \qa).  Combining both \pointclouds and RGB provides improvements to both obstacle avoidance and leveraging semantic information. %

%% file: sections/main/conclusion.tex
\vspace{\sectionReduceTop}
\section{Conclusion}
\vspace{\sectionReduceBot}

We present an extension of the task of \eqa to photorealistic environments utilizing the Matterport 3D dataset and propose the \mpeqa v1 dataset.  We then present a thorough study of 2 navigation baselines and 2 different navigation architectures with 8 different input variations. We develop an end-to-end trainable navigation model capable of learning goal-driving navigation policies directly from 3D \pointclouds.  We provide analysis and insight into the factors that affect navigation performance and propose a novel weighting scheme -- \textit{Inflection Weighting} -- that increases the effectiveness of behavior cloning.  We demonstrate that two the navigation baselines, random and forward-only, are quite strong under the evaluation settings presented by \cite{embodiedqa}.  Our work serves as a step towards bridging the gap between  \textbf{\emph{internet vision}}-style problems and the goal of \textbf{\emph{vision for embodied perception}}. 

%% file: sections/supp/supp_content.tex
\input{sections/supp/supp_body.tex}

\clearpage
\input{sections/supp/main_tab_v2.tex}

\clearpage
\input{sections/supp/noiw_tab_v2.tex}

\clearpage

\input{sections/supp/main_tab_cis.tex}

\clearpage

\input{sections/supp/noiw_tab_cis.tex}

%% file: sections/supp/supp_body.tex
\vspace{\sectionReduceTop}
\section{Color Label Collection Interface}
\vspace{\sectionReduceBot}

\begin{figure*}
    \centering
    \includegraphics[width=0.9\textwidth, trim={0 0 0 0.1cm},clip]{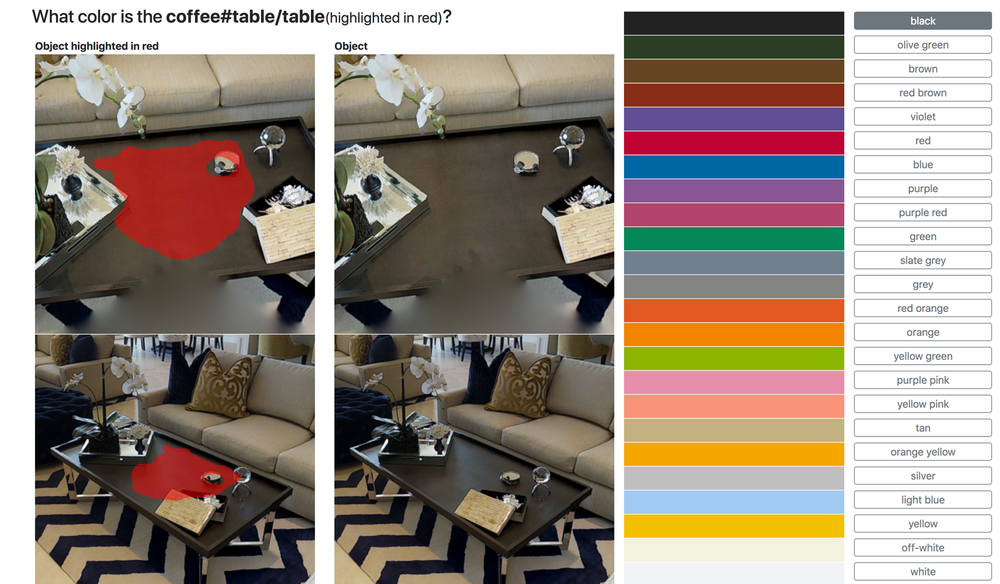}
    \caption{Interface shown to AMT workers for collecting the dominate color name for objects in the Matterport3D dataset~\cite{Matterport3D}.  Workers were shown up to two good views of the object (some objects only had one good view) and the corresponding instance segmentation mask and asked to selected the dominate color of the object from a predefined list of colors.}
    \label{fig:amt}
    \vspace{\captionReduceBot}
\end{figure*}

\reffig{fig:amt} shows the interface we used to collect dominate color annotations from workers on Amazon Mechanical Turk.

\vspace{\sectionReduceBot}
\section{Point Cloud Rendering}
\vspace{\sectionReduceTop}

In order to render 2.5D \rgbd frames, we
first construct a global point cloud from all of the panoramas
provided in an environment in the Matterport3D dataset~\cite{Matterport3D}.  Next, we determine what parts of the global point cloud are
in the agent's current view.  The agent's current position, $A_{pos}$, camera parameters (field of view, and
aspect ratio), and the mesh reconstruction are used to determine which points are within its view pyramid (frustum).

For each point $p_i$ in view, we first check if the point lies within the agents view frustum using the extrinsic and intrinsic camera matrices. After determining which points lie within the view frustum, we check for occlusions. We then draw a line between the agent's camera and $p_i$, let $L_i$ be that line. We intersect this line with the provided mesh and keep the intersection that is \textit{closest} to the agent, let $ \argmin(L_i \cap \mathcal{M})$ be that intersection. If the distance to the \textit{closest} intersection with the mesh, $||A_{pos} - \argmin(L_i \cap \mathcal{M})||$, is less than the distance to the point, $||A_{pos} - p_i||$, indicating that there is a closer (occluding) point, we remove the point.  We also perform this check in the other direction and remove the point if $||A_{pos} - p_i || < ||A_{pos} - \argmin(L_i \cap \mathcal{M})||$, indicating that the point is in the free-space. Points in the free-space are a result of panorama alignment errors and scanning oddities from reflections. In practice, we find that equality is too strict of a criteria due to mesh reconstruction errors, we instead remove a point if the absolute difference of the distances is greater than $\epsilon = 0.25~\text{cm}$, $\text{abs}(||A_{pos} - \argmin(L_i \cap \mathcal{M})|| -  ||A_{pos} - p_i||) > \epsilon$.

The implementation implied by the description above would be very slow.  What we have described above implies something akin to a ray-tracing rendering.  As in normal graphics pipelines, we can significantly speed this up by approximation with rasterization.  We rasterize the provided mesh at $A_{pos}$ and capture the depth buffer.  The depth buffer provides the distance from the agent to the \textit{closest} intersection with the mesh for some finite number of rays.  Let $R_j$ be a ray in that set and  $||A_{pos} - \argmin(R_j \cap \mathcal{M})||$ be the distance to the closest intersection with the mesh as provided by the depth buffer. We can then approximate $||A_{pos} - \argmin(L_i \cap \mathcal{M})||$ for every point $p_i$ in the point cloud by finding the $R_j$ that is the closest to parallel to $L_i$ and using the value of $||A_{pos} - \argmin(R_j \cap \mathcal{M})||$.

Further, a single global point cloud in Matterport3D environments has hundreds of millions of points and, in theory, we would need to check every single one to determine what points are visible.  This would be quite slow and use a prohibitive amount of memory.  We alleviate this by creating a significantly sparser point cloud and perform an initial visibility check on that instead.  We then recheck the dense point cloud only in areas of the sparse point cloud that passed the initial visibility check.

\vspace{\sectionReduceTop}
\section{Perception Models}
\vspace{\sectionReduceBot}
Here we provide full details on the architectures of our perceptual encoders and the decoder heads used for their pre-training tasks.

\vspace{\subsectionReduceTop}
\subsection{PC -- PointNet++}
\vspace{\subsectionReduceBot}

We use notation similar to the notation from Qi \etal \cite{qi2017pointnet++} to specify our PointNet++ architecture.  
Let $P^{(0)}$ be the input \pointcloud.
Then, 
\begin{align*}
    &SA^{(k+1)}(P^{(k)}, N, [r^{(1)},...,r^{(m)}], \\
    &~~~[[l_1^{(1)}, ..., l_d^{(1)}],...,[l_1^{(m)}, ..., l_d^{(m)}]])
\end{align*}
\noindent
is a set abstraction module with multi-scale grouping that takes input the $k~th$ level point cloud, $P^{(k)}$, and produces the $k+1~th$ level point cloud,  $P^{(k + 1)}$, with $N$ points.  See section 4.1 in the main paper for a full description of a set abstraction module with single-scale grouping. A set abstraction module with multi-scale grouping is a logical extension. The feature descriptor for each point in $P^{(k+1)}$ is calculated across $m$ different scales ($m$ balls with different radii). $[l_1^{(1)}, ..., l_d^{(1)}]$ specifies the number of output channels for each layer in the shared-weighted multi-layer perceptron (MLP) used a each scale. The feature descriptor for any point is then the concatenation of the feature descriptor calculated at each scale.

As an analogy, a multi-scale convolution would be achieved by by first convolving the input with convolutions of different kernel sizes and then concatenating the outputs from each convolution.

$SA^{(k+1)}(P^{(k)}, [l_1, ..., l_d])$ is a global set abstraction module and produces a $l_d$ dimensional feature vector.

Our encoder is specified by
\begin{align*}
    & SA^{(1)}(P^{(0)}, 1024, [0.05, 0.1], \\ 
        &~~~ [[32, 32, 64], [32, 64, 128]]) \\
    & SA^{(2)}(P^{(1)}, 256, [0.1, 0.2, 0.4], \\
        &~~~ [[64, 128, 128], [128, 128, 256], \\
        &~~~ [128, 128, 256, 256]]) \\
    & SA^{(3)}(P^{(2)}, 64, [0.4, 0.8], \\
        &~~~ [[128, 128, 128, 256, 256], \\
        &~~~ [128, 128, 256, 256, 256, 512]]) \\
    & SA^{(4)}(P^{(3)}, [256, 512, 1024])
\end{align*}

$FP(P^{(k+1}, P^{(k)}, use\_skip, [l_1,...l_d])$ is a feature propagation layer that transfers the features from $P^{(k+1)}$ to $P^{(k)}$ \cite{qi2017pointnet++}.  For each point in $P^{(k)}$, its three nearest neighbors in $P^{(k+1)}$ are found and their feature descriptors are combined via inverse-distance weighted interpolation.  The interpolated feature descriptor is then processed by shared-weight MLP with output channels specified by $[l_1,...l_d]$.
$use\_skip$ is a $True$ or $False$ boolean that determines if any existing features of $P^{(k)}$ are used. If $use\_skip$ is true, the the interpolated feature and the existing feature are concatenated together before the shared-weight MLP. If $use\_skip$ is false, only the interpolated feature is processed by the shared-weight MLP.

\noindent
The semantic segmentation head is specified as
\begin{align*}
    Prop & = FP(P^{(4)}, P^{(3)}, True, [256, 512]) \\
    Prop & = FP(Prop, P^{(2)}, True, [256, 256]) \\
    Prop & = FP(Prop, P^{(1)}, True, [256, 256]) \\
    Output& = FP(Prop, P^{(0)}, True, [256, 128, 128, K]) \\
\end{align*}
\noindent
where $K$ is the number of class, $40$ in our case.

\noindent
The color prediction head is specified as
\begin{align*}
    Prop & = FP(P^{(4)}, P^{(3)}, True, [256, 512]) \\
    Prop & = FP(Prop, P^{(2)}, True, [256, 256]) \\
    Prop & = FP(Prop, P^{(1)}, False, [256, 256]) \\
    Output& = FP(Prop, P^{(0)}, False, [256, 128, 128, 3]) \\
\end{align*}

\noindent
The structure prediction head is specified as
\begin{align*}
    (a)& ~ FC(P^{(4)}, 256) \\
    (b)& ~ FC((a), 256) \\
    (c)& ~ FC((b), N*3) \\
\end{align*}
$FC(in, d_{out})$ is a fully connected layer that takes $in$ and produces a vector of size $d_{out}$.  The output of the structure decoding head, $(c)$, is then reshaped into a $(N,3)$ point cloud.

\xhdr{Training details.}
We train with a batch size of $32$ and utilize Adam \cite{kingma_iclr15}.  We use an initial learning rate of $10^{-3}$ and decay the learning rate by $70\%$ every $6.2 \times 10^{-3}$ batches.  We select the checkpoint by performance on a held-out validation set.
 
\vspace{\subsectionReduceTop}
\subsection{RGB -- ResNet50}
\vspace{\subsectionReduceBot}

We use the initial convolution and $4$ residual blocks from ResNet50~\cite{he2016deep} as the encoder for RGB images.  Each of the three decoders (semantic, depth, and color) are identical and consistent of 1x1 convolutions and bi-linear interpolation.

Let $RB2$, $RB3$, and $RB4$ be the outputs from the second, third, and fourth residual blocks in ResNet50 respectively.  Each decoder is then parameterized as

\begin{align*}
    Up4& = 1x1Conv(RB4, C) \\
    Up3& = UpSample(Up4) + 1x1Conv(RB3, C) \\ 
    Up2& = UpSample(Up3) + 1x1Conv(RB2, C) \\
    Output& = UpSample(Up2)
\end{align*}

\noindent
$1x1Conv$ is simply a 1-by-1 convectional layer that transforms its input to have $C$ channels.  $UpSample$ is a bi-linear interpolation layer that up-samples its input to be the necessary size.
Where $C$ is the number of output channels for the given decoder type, $K$ for semantic, $1$ for depth, and $3$ for color.

\xhdr{Training details.}
Due to the high prevalence of walls and floors among indoor scenes, we use class weighted cross-entropy loss for semantic segmentation. For depth and autoencoding, we use smooth-$\ell_1$  The total loss is then the weighted sum of the individual task losses:
\begin{align*}
L = \lambda_{Seg.} L_{Seg.} + \lambda_{Depth} L_{Depth} + \lambda_{AE} L_{AE}
\end{align*}

\noindent
We find $\lambda_{Seg.} = 0.1$, $\lambda_{Depth} = 10$, and $\lambda_{AE} = 10$ balances the magnitudes of the various losses and works well.

\reftab{tab:rgb_results} provides a comparison of pre-training tasks when trained for separately vs. jointly and comparison with the Shallow CNN used in Das \etal \cite{embodiedqa}. The experiments showed that training jointly doesn't sacrifice performance on depth and autoencoding.

We train with a batch size of $20$ and utilize Adam \cite{kingma_iclr15}.  We use an initial learning rate of $10^{-5}$, maximum epochs considered $300$, we sampled every $3$-rd frame from shortest paths. The checkpoint was selected by performance on a held-out validation set.

\begin{table*}[t!]
\centering
\resizebox{0.95\textwidth}{!}{
\begin{tabular}[t]{c l c c c c c c c}
\toprule
    & Model & Total loss & AE loss & Depth loss & Sem. loss & PA & MPA & MIOU \\
\midrule
    & Shallow CNN \cite{embodiedqa} -- All Tasks & 0.369 & 0.0047 & 0.0077 & 2.45  & 0.384 & 0.14 & 0.070 \\
    & ResNet50 -- AE only & 1.774 & \textbf{0.0030} & 0.0873 & 8.71 & 0.008 & 0.02 & 0.002 \\
    &ResNet50 -- Depth only & 2.311 & 0.1051&0.0072 & 11.89& 0.005 & 0.02 & 0.002 \\
    &ResNet50 -- AE and depth only & 1.003 & 0.0034 & \textbf{0.0067} & 9.02 & 0.006 & 0.03 & 0.002 \\
    &ResNet50 -- All Tasks & \textbf{0.356} & 0.0040 & 0.0069 & \textbf{2.48} & \textbf{0.390} & \textbf{0.15} & \textbf{0.078} \\
\bottomrule
\end{tabular}}
\vspace{-5pt}
\caption{Performance on the \mpeqa v1 validation set for Shallow CNN~\cite{embodiedqa} and for ResNet50 when trained for autoencdoing only, for depth only, for autoencoding and depth, and for all tasks jointly. For segmentation we report the overall pixel accuracy (PA), mean pixel
accuracy (MPA) averaged over all semantic classes and the mean IOU intersection over union (MIOU). For depth and autoencoder, we
report the smooth-$\ell_1$ on the validation set.}
\label{tab:rgb_results}
\vspace{\captionReduceBot}
\end{table*}

\vspace{\sectionReduceTop}
\section{Question answering model training}
\vspace{\sectionReduceBot}

\begin{compactenum}[\hspace{5pt}--]
\item \texttt{lstm-question-only}, we train with a batch size of 40 and utilize Adam with a learning rate of $10^{-3}$;
\item \texttt{nn-question-only} and \texttt{bow-question-only} we use the code provided in \cite{anand2018blindfold};

\item \texttt{attention+*}, we train with a batch size of 20 and utilize Adam with a learning rate of $10^{-3}$;
\item \texttt{spatial+RGB+Q} 
, we train with a batch size of 32 and utilize Adam with a learning rate of $10^{-3}$
\end{compactenum}

For all models, the best checkpoint is selected via performance on a held out validation set.

\vspace{\sectionReduceTop}
\section{Navigation model training}
\vspace{\sectionReduceBot}

For models trained with and without inflection weighted loss, we train and select checkpoints with the same procedure.

\begin{compactenum}[\hspace{5pt}$\cdot$]
\item \texttt{R+*}, we train with a batch size of 20 and utilize Adam with a learning rate of $10^{-3}$;
\item \texttt{M+*}, we train with a batch size of 5 full sequences and utilize Adam with a learning rate of $2 \times 10^{-4}$.  Note that due to the length of sequences, we compute the forward and backward pass for each sequence individually and average the gradients
\end{compactenum}

Due to the difference in end-to-end evaluation and teacher-forcing accuracy, we use the following method to select checkpoints for navigation models:
First we run non-minimal suppression with a window size of 5 on teacher forcing validation loss (vanilla cross-entropy for \texttt{NoIW-*} models, and inflection weighted cross-entropy otherwise).  After NMS, we select the top 5 checkpoints by validation loss and run them through end-to-end evaluation on the validation set.  The best checkpoint is the model that has the highest value of $\mathbf{QA} + \mathbf{d_{\Delta}}$ at $T_{-50}$.  We find that inflection weighted cross-entropy is a significantly better predictor of end-to-end performance than vanilla cross-entropy.  Interestingly, we also find that teacher forcing validation accuracy is a good predictor of end-to-end performance when training with inflection weighting and don't see a significant difference in performance if the procedure above is run with  teacher forcing validation error instead of loss.

\vspace{\sectionReduceTop}
\section{Results}
\vspace{\sectionReduceBot}

We provide the full tables we first analyzed in their entirety and then sliced for the analysis provided in the main paper. \reftab{tab:main_results} shows our primary results with inflection weighting and \reftab{tab:no_iw} shows the same set of models trained without inflection weighting.

We also provide the full tables with confidence intervals.  Confidence intervals are 90\% confidence intervals calculated with empirical bootstrapping.  See \reftab{tab:main_results} and \reftab{tab:no_iw_cis}.

As a reminder, we use the following notation to specify our models:  For the base architecture, \texttt{R} denotes reactive models and \texttt{M} denotes memory models.  The base architectures are then augmented with their input types, \texttt{+PC}, \texttt{+RGB}, and \texttt{+Q}. So a memory model that utilizes \pointclouds (but no question) is denoted as \texttt{M+PC}. Unless otherwise specified (by the prefix \texttt{NoIW}), models are trained with inflection weighting.
We denote the two baseline navigators, forward-only and random, as \texttt{Fwd} and \texttt{Random}, respectively.

\vspace{\subsectionReduceTop}
\subsection{Navigation Performance Correlation Analysis}
\vspace{\subsectionReduceBot}

\begin{figure}
    \centering
    \includegraphics[width=\columnwidth]{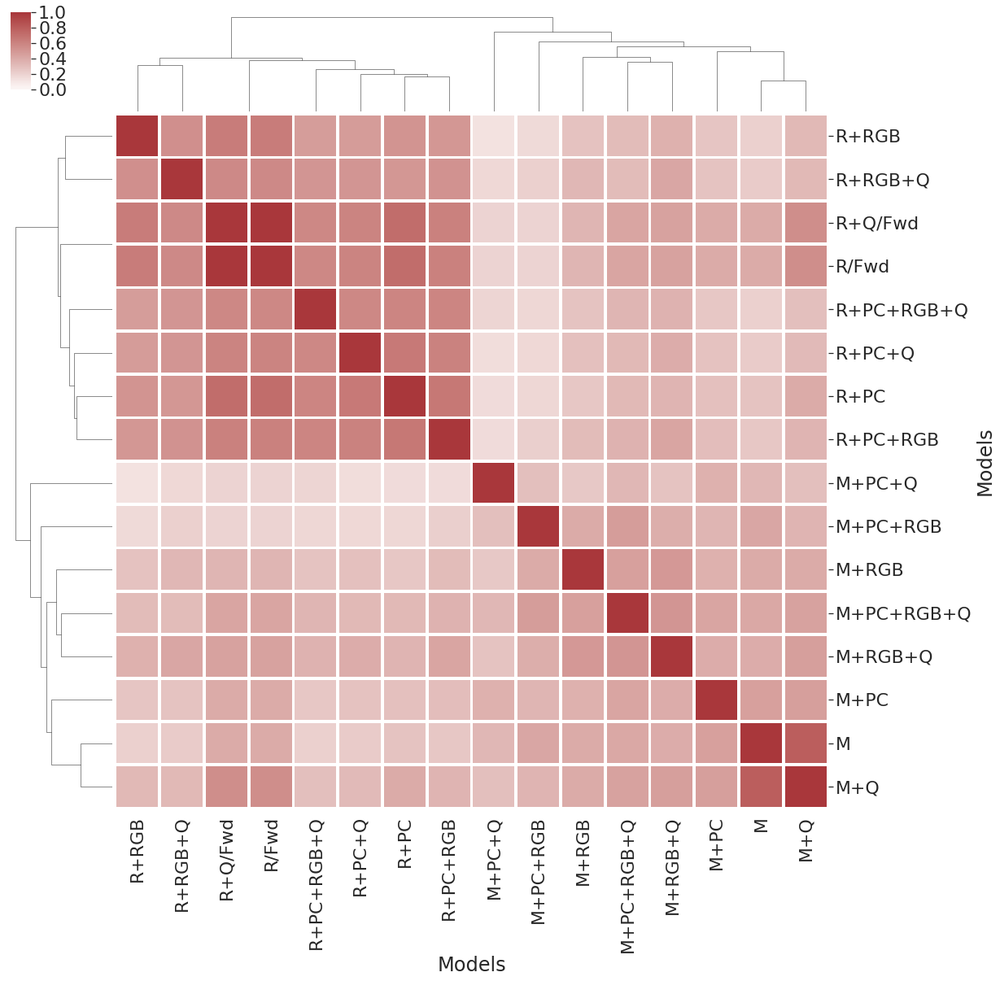}
    \caption{Correlation and clustering of correlation for $d_T$ per episode at $T_{-30}$. Reactive models build a strong cluster, while memory models are less correlated. Usage of the question changes navigators with vision significantly.
    }
    \label{fig:d_t_corr}
    \vspace{\captionReduceBot}
\end{figure}

Here we provide some additional insight on the effect the question has on our navigation models.  We find that while the question does significantly impact performance on average, it does significantly change a model's behavior.

\reffig{fig:d_t_corr} shows the Pearson correlation correlation coefficient between $\mathbf{d_T}$ at $T_{-30}$ for all models and cluster discovery using the Nearest Point Algorithm.  All navigators that use the reactive base architecture build a strong cluster and are highly correlated with each-other.  
The vision-less memory models are also well correlated with each-other, indicating that the question has little relation to the distribution of actions along a shortest path. Memory models that use vision don't tend to be well correlated with their counterpart that also uses the question in-spite of the question having little effect on overall navigation metrics.  

%% file: sections/supp/main_tab_v2.tex
\begin{table*}[t!]
\setlength\tabcolsep{3pt}
\renewcommand{\arraystretch}{1.2}
\resizebox{\textwidth}{!}{
\begin{tabular}{l l l l c c c c c c l c c c l c c c l c c c l c c c l c c c }
& &  \multicolumn{24}{c}{\textbf{Navigation}} &~~~ & \multicolumn{3}{c}{\textbf{QA}}\\
\cmidrule{4-26}\cmidrule{28-30}
&\textbf{Navigator}&& \multicolumn{3}{c}{$\mathbf{d_0}$ {\scriptsize(For reference)}} &
& \multicolumn{3}{c}{$\mathbf{d_T}$ {\scriptsize(Lower is better)}} &
& \multicolumn{3}{c}{$\mathbf{d_{min}}$ {\scriptsize(Lower is better)}} &
& \multicolumn{3}{c}{$\mathbf{d_{\Delta}}$ {\scriptsize(Higher is better)}} &
& \multicolumn{3}{c}{$\mathbf{\%_{collision}}$ {\scriptsize(Lower is better)}} &
& \multicolumn{3}{c}{$\mathbf{IoU_{T}}$ {\scriptsize(Higher is better)}} &
& \multicolumn{3}{c}{$\mathbf{Top-1}$ {\scriptsize(Higher is better)}}\\
& && \scriptsize$T_{-10}$ & \scriptsize$T_{-30}$ & \scriptsize$T_{-50}$ &
& \scriptsize$T_{-10}$ & \scriptsize$T_{-30}$ & \scriptsize$T_{-50}$ &
& \scriptsize$T_{-10}$ & \scriptsize$T_{-30}$ & \scriptsize$T_{-50}$ &
& \scriptsize$T_{-10}$ & \scriptsize$T_{-30}$ & \scriptsize$T_{-50}$ &
 & \scriptsize$T_{-10}$ & \scriptsize$T_{-30}$ & \scriptsize$T_{-50}$ &
 & \scriptsize$T_{-10}$ & \scriptsize$T_{-30}$ & \scriptsize$T_{-50}$ &
 & \scriptsize$T_{-10}$ & \scriptsize$T_{-30}$ & \scriptsize$T_{-50}$ \\
 \toprule
&\texttt{ R }&& $0.354$ & $1.898$ & $3.547$ && $0.933$ & $1.330$ & $2.154$ && $\mathbf{0.011}$ & $0.346$ & $1.397$ && $-0.579$ & $0.568$ & $1.393$ && $79.554$ & $66.182$ & $62.563$ && $0.062$ & $0.050$ & $0.030$ && $0.390$ & $0.379$ & $0.354$ \\
&\texttt{ R+Q }&& $0.354$ & $1.898$ & $3.547$ && $0.933$ & $1.330$ & $2.154$ && $\mathbf{0.011}$ & $0.346$ & $1.397$ && $-0.579$ & $0.568$ & $1.393$ && $79.554$ & $66.182$ & $62.563$ && $0.062$ & $0.050$ & $0.030$ && $0.390$ & $0.379$ & $0.354$ \\
\\ 
&\texttt{ R+RGB }&& $0.354$ & $1.898$ & $3.547$ && $1.194$ & $1.617$ & $2.340$ && $0.040$ & $0.375$ & $1.349$ && $-0.840$ & $0.281$ & $1.207$ && $59.959$ & $51.460$ & $48.425$ && $0.077$ & $0.058$ & $0.031$ && $0.395$ & $0.396$ & $0.372$ \\
&\texttt{ R+RGB+Q }&& $0.354$ & $1.898$ & $3.547$ && $1.407$ & $1.740$ & $2.521$ && $0.034$ & $0.340$ & $1.332$ && $-1.053$ & $0.157$ & $1.026$ && $51.128$ & $44.160$ & $42.692$ && $0.111$ & $0.070$ & $0.054$ && $0.383$ & $0.388$ & $0.375$ \\
\\ 
&\texttt{ R+PC }&& $0.354$ & $1.898$ & $3.547$ && $1.428$ & $1.754$ & $2.352$ && $0.021$ & $\mathbf{0.320}$ & $1.164$ && $-1.074$ & $0.144$ & $1.195$ && $50.148$ & $41.612$ & $42.203$ && $0.070$ & $0.067$ & $0.047$ && $0.356$ & $0.394$ & $0.375$ \\
&\texttt{ R+PC+Q }&& $0.354$ & $1.898$ & $3.547$ && $1.514$ & $1.812$ & $2.394$ && $0.033$ & $0.325$ & $\mathbf{1.160}$ && $-1.160$ & $0.085$ & $1.153$ && $46.910$ & $36.303$ & $39.012$ && $0.059$ & $0.052$ & $0.043$ && $0.364$ & $0.364$ & $0.363$ \\
\\ 
&\texttt{ R+PC+RGB }&& $0.354$ & $1.898$ & $3.547$ && $1.547$ & $1.791$ & $2.336$ && $0.020$ & $0.322$ & $1.211$ && $-1.193$ & $0.107$ & $1.211$ && $44.941$ & $34.859$ & $37.138$ && $0.084$ & $0.077$ & $0.044$ && $0.374$ & $0.390$ & $0.366$ \\
&\texttt{ R+PC+RGB+Q }&& $0.354$ & $1.898$ & $3.547$ && $1.539$ & $1.843$ & $2.420$ && $0.032$ & $0.323$ & $1.170$ && $-1.185$ & $0.055$ & $1.127$ && $42.018$ & $34.318$ & $37.069$ && $0.067$ & $0.072$ & $0.055$ && $0.370$ & $0.395$ & $0.369$ \\

\cmidrule{2-30}
&\texttt{ M }&& $0.354$ & $1.898$ & $3.547$ && $\mathbf{0.366}$ & $\mathbf{0.830}$ & $1.833$ && $0.090$ & $0.505$ & $1.460$ && $\mathbf{-0.012}$ & $\mathbf{1.068}$ & $1.714$ && $6.903$ & $10.989$ & $23.250$ && $0.128$ & $0.091$ & $0.081$ && $0.365$ & $0.375$ & $0.363$ \\
&\texttt{ M+Q }&& $0.354$ & $1.898$ & $3.547$ && $0.508$ & $0.933$ & $1.920$ && $0.052$ & $0.426$ & $1.421$ && $-0.154$ & $0.965$ & $1.627$ && $16.268$ & $19.808$ & $32.856$ && $0.147$ & $0.109$ & $0.068$ && $0.391$ & $0.395$ & $0.376$ \\
\\ 
&\texttt{ M+RGB }&& $0.354$ & $1.898$ & $3.547$ && $0.637$ & $1.157$ & $2.177$ && $0.099$ & $0.538$ & $1.479$ && $-0.283$ & $0.741$ & $1.370$ && $12.582$ & $15.130$ & $26.179$ && $0.188$ & $0.136$ & $0.075$ && $0.397$ & $0.403$ & $0.384$ \\
&\texttt{ M+RGB+Q }&& $0.354$ & $1.898$ & $3.547$ && $0.707$ & $1.171$ & $2.194$ && $0.071$ & $0.423$ & $1.386$ && $-0.353$ & $0.727$ & $1.353$ && $14.212$ & $15.908$ & $25.578$ && $0.189$ & $0.141$ & $0.083$ && $\mathbf{0.407}$ & $0.394$ & $0.384$ \\
\\ 
&\texttt{ M+PC }&& $0.354$ & $1.898$ & $3.547$ && $0.494$ & $1.020$ & $1.817$ && $0.098$ & $0.484$ & $1.236$ && $-0.140$ & $0.878$ & $1.730$ && $6.647$ & $9.169$ & $18.319$ && $0.163$ & $0.114$ & $0.083$ && $0.396$ & $\mathbf{0.411}$ & $\mathbf{0.390}$ \\
&\texttt{ M+PC+Q }&& $0.354$ & $1.898$ & $3.547$ && $0.502$ & $1.030$ & $1.910$ && $0.081$ & $0.497$ & $1.272$ && $-0.148$ & $0.868$ & $1.637$ && $5.584$ & $\mathbf{8.833}$ & $\mathbf{15.783}$ && $0.184$ & $0.158$ & $\mathbf{0.118}$ && $0.382$ & $0.387$ & $0.374$ \\
\\ 
&\texttt{ M+PC+RGB }&& $0.354$ & $1.898$ & $3.547$ && $0.461$ & $0.940$ & $\mathbf{1.791}$ && $0.103$ & $0.513$ & $1.269$ && $-0.107$ & $0.958$ & $\mathbf{1.756}$ && $\mathbf{4.957}$ & $9.574$ & $18.890$ && $\mathbf{0.209}$ & $\mathbf{0.179}$ & $0.111$ && $0.381$ & $0.393$ & $0.363$ \\
&\texttt{ M+PC+RGB+Q }&& $0.354$ & $1.898$ & $3.547$ && $0.574$ & $1.044$ & $1.898$ && $0.083$ & $0.431$ & $1.203$ && $-0.220$ & $0.854$ & $1.649$ && $8.328$ & $10.674$ & $19.797$ && $\mathbf{0.209}$ & $0.148$ & $0.112$ && $0.389$ & $0.390$ & $0.373$ \\

\cmidrule{2-30}
&\texttt{ Random }&& $0.354$ & $1.898$ & $3.547$ && $0.912$ & $1.273$ & $2.654$ && $0.048$ & $0.796$ & $2.263$ && $-0.558$ & $0.625$ & $0.893$ && $13.775$ & $10.708$ & $10.677$ && $0.098$ & $0.072$ & $0.041$ && $0.365$ & $0.368$ & $0.364$ \\
&\texttt{ ShortestPath }&& $0.354$ & $1.898$ & $3.547$ && $0.005$ & $0.005$ & $0.005$ && $0.005$ & $0.005$ & $0.005$ && $0.349$ & $1.893$ & $3.542$ && $0.000$ & $0.000$ & $0.000$ && $0.581$ & $0.581$ & $0.581$ && $0.451$ & $0.451$ & $0.451$ \\
\bottomrule
\end{tabular}

}\\[3pt]
\caption{Evaluation of \eqa agents trained with inflection weighting on navigation and answering metrics for the \mpeqa v1 test set.
RGB models perceive the world via RGB images and use ResNet50.  PC models perceive the world via point clouds and use PointNet++.  PC+RGB models use both perception modalities and their respective networks.}
\label{tab:main_results}
\end{table*}

%% file: sections/supp/noiw_tab_v2.tex
\begin{table*}[t!]
\setlength\tabcolsep{3pt}
\renewcommand{\arraystretch}{1.2}
\resizebox{\textwidth}{!}{
\begin{tabular}{l l l l c c c c c c l c c c l c c c l c c c l c c c l c c c }
& &  \multicolumn{24}{c}{\textbf{Navigation}} &~~~ & \multicolumn{3}{c}{\textbf{QA}}\\
\cmidrule{4-26}\cmidrule{28-30}
&\textbf{Navigator}&& \multicolumn{3}{c}{$\mathbf{d_0}$ {\scriptsize(For reference)}} &
& \multicolumn{3}{c}{$\mathbf{d_T}$ {\scriptsize(Lower is better)}} &
& \multicolumn{3}{c}{$\mathbf{d_{min}}$ {\scriptsize(Lower is better)}} &
& \multicolumn{3}{c}{$\mathbf{d_{\Delta}}$ {\scriptsize(Higher is better)}} &
& \multicolumn{3}{c}{$\mathbf{\%_{collision}}$ {\scriptsize(Lower is better)}} &
& \multicolumn{3}{c}{$\mathbf{IoU_{T}}$ {\scriptsize(Higher is better)}} &
& \multicolumn{3}{c}{$\mathbf{Top-1}$ {\scriptsize(Higher is better)}}\\
& && \scriptsize$T_{-10}$ & \scriptsize$T_{-30}$ & \scriptsize$T_{-50}$ &
& \scriptsize$T_{-10}$ & \scriptsize$T_{-30}$ & \scriptsize$T_{-50}$ &
& \scriptsize$T_{-10}$ & \scriptsize$T_{-30}$ & \scriptsize$T_{-50}$ &
& \scriptsize$T_{-10}$ & \scriptsize$T_{-30}$ & \scriptsize$T_{-50}$ &
 & \scriptsize$T_{-10}$ & \scriptsize$T_{-30}$ & \scriptsize$T_{-50}$ &
 & \scriptsize$T_{-10}$ & \scriptsize$T_{-30}$ & \scriptsize$T_{-50}$ &
 & \scriptsize$T_{-10}$ & \scriptsize$T_{-30}$ & \scriptsize$T_{-50}$ \\
 \toprule
&\texttt{ NoIW-R }&& $0.354$ & $1.898$ & $3.547$ && $0.933$ & $1.330$ & $\mathbf{2.154}$ && $\mathbf{0.011}$ & $0.346$ & $1.397$ && $-0.579$ & $0.568$ & $\mathbf{1.393}$ && $79.554$ & $66.182$ & $62.563$ && $0.062$ & $0.050$ & $0.030$ && $0.390$ & $0.379$ & $0.354$ \\
&\texttt{ NoIW-R+Q }&& $0.354$ & $1.898$ & $3.547$ && $0.933$ & $1.330$ & $\mathbf{2.154}$ && $\mathbf{0.011}$ & $0.346$ & $1.397$ && $-0.579$ & $0.568$ & $\mathbf{1.393}$ && $79.554$ & $66.182$ & $62.563$ && $0.062$ & $0.050$ & $0.030$ && $0.390$ & $0.379$ & $0.354$ \\
\\ 
&\texttt{ NoIW-R+RGB }&& $0.354$ & $1.898$ & $3.547$ && $1.419$ & $1.713$ & $2.528$ && $0.041$ & $0.404$ & $1.417$ && $-1.065$ & $0.185$ & $1.019$ && $56.718$ & $50.376$ & $45.866$ && $0.086$ & $0.049$ & $0.035$ && $0.382$ & $0.386$ & $0.360$ \\
&\texttt{ NoIW-R+RGB+Q }&& $0.354$ & $1.898$ & $3.547$ && $1.405$ & $1.829$ & $2.658$ && $0.051$ & $0.455$ & $1.463$ && $-1.051$ & $0.069$ & $0.889$ && $43.226$ & $38.538$ & $36.172$ && $0.075$ & $0.060$ & $0.054$ && $0.384$ & $0.390$ & $0.373$ \\
\\ 
&\texttt{ NoIW-R+PC }&& $0.354$ & $1.898$ & $3.547$ && $1.385$ & $1.662$ & $2.483$ && $0.026$ & $\mathbf{0.343}$ & $1.294$ && $-1.031$ & $0.236$ & $1.064$ && $43.067$ & $34.100$ & $37.078$ && $0.074$ & $0.061$ & $0.043$ && $0.375$ & $0.398$ & $0.376$ \\
&\texttt{ NoIW-R+PC+Q }&& $0.354$ & $1.898$ & $3.547$ && $1.515$ & $1.858$ & $2.646$ && $0.038$ & $0.394$ & $1.333$ && $-1.161$ & $0.040$ & $0.901$ && $37.669$ & $31.714$ & $\mathbf{33.563}$ && $0.078$ & $0.059$ & $0.054$ && $0.372$ & $\mathbf{0.399}$ & $0.378$ \\
\\ 
&\texttt{ NoIW-R+PC+RGB }&& $0.354$ & $1.898$ & $3.547$ && $1.462$ & $1.759$ & $2.523$ && $0.025$ & $0.347$ & $\mathbf{1.285}$ && $-1.108$ & $0.139$ & $1.024$ && $53.785$ & $41.955$ & $40.293$ && $0.067$ & $0.052$ & $0.042$ && $0.384$ & $0.389$ & $0.369$ \\
&\texttt{ NoIW-R+PC+RGB+Q }&& $0.354$ & $1.898$ & $3.547$ && $1.297$ & $1.704$ & $2.543$ && $0.030$ & $0.425$ & $1.417$ && $-0.943$ & $0.194$ & $1.004$ && $47.506$ & $39.554$ & $36.242$ && $0.069$ & $0.062$ & $0.046$ && $0.368$ & $0.375$ & $0.353$ \\

\cmidrule{2-30}
&\texttt{ NoIW-M }&& $0.354$ & $1.898$ & $3.547$ && $0.933$ & $1.330$ & $2.186$ && $\mathbf{0.011}$ & $0.346$ & $1.430$ && $-0.579$ & $0.568$ & $1.361$ && $79.554$ & $66.182$ & $62.818$ && $0.062$ & $0.050$ & $0.029$ && $0.390$ & $0.379$ & $0.356$ \\
&\texttt{ NoIW-M+Q }&& $0.354$ & $1.898$ & $3.547$ && $0.933$ & $1.330$ & $2.202$ && $\mathbf{0.011}$ & $0.346$ & $1.445$ && $-0.579$ & $0.568$ & $1.345$ && $79.554$ & $66.182$ & $62.931$ && $0.062$ & $0.050$ & $0.029$ && $0.390$ & $0.379$ & $0.354$ \\
\\ 
&\texttt{ NoIW-M+RGB }&& $0.354$ & $1.898$ & $3.547$ && $0.902$ & $1.396$ & $2.512$ && $0.024$ & $0.400$ & $1.608$ && $-0.548$ & $0.502$ & $1.035$ && $67.347$ & $59.661$ & $59.378$ && $0.080$ & $0.061$ & $0.033$ && $\mathbf{0.406}$ & $0.384$ & $0.353$ \\
&\texttt{ NoIW-M+RGB+Q }&& $0.354$ & $1.898$ & $3.547$ && $0.911$ & $1.394$ & $2.573$ && $0.024$ & $0.410$ & $1.644$ && $-0.557$ & $0.504$ & $0.974$ && $66.198$ & $59.317$ & $58.941$ && $0.094$ & $0.064$ & $0.030$ && $0.390$ & $0.380$ & $0.356$ \\
\\ 
&\texttt{ NoIW-M+PC }&& $0.354$ & $1.898$ & $3.547$ && $0.811$ & $1.245$ & $2.244$ && $0.034$ & $0.370$ & $1.414$ && $-0.457$ & $0.652$ & $1.303$ && $37.865$ & $30.964$ & $38.521$ && $0.113$ & $0.114$ & $\mathbf{0.091}$ && $0.386$ & $\mathbf{0.399}$ & $0.376$ \\
&\texttt{ NoIW-M+PC+Q }&& $0.354$ & $1.898$ & $3.547$ && $\mathbf{0.790}$ & $\mathbf{1.233}$ & $2.213$ && $0.038$ & $0.379$ & $1.416$ && $\mathbf{-0.436}$ & $\mathbf{0.665}$ & $1.334$ && $36.215$ & $31.431$ & $38.911$ && $0.115$ & $0.099$ & $0.076$ && $0.393$ & $\mathbf{0.399}$ & $\mathbf{0.379}$ \\
\\ 
&\texttt{ NoIW-M+PC+RGB }&& $0.354$ & $1.898$ & $3.547$ && $0.929$ & $1.405$ & $2.435$ && $0.016$ & $0.375$ & $1.526$ && $-0.575$ & $0.492$ & $1.112$ && $61.658$ & $51.595$ & $52.093$ && $0.100$ & $0.075$ & $0.041$ && $0.388$ & $0.385$ & $0.353$ \\
&\texttt{ NoIW-M+PC+RGB+Q }&& $0.354$ & $1.898$ & $3.547$ && $0.871$ & $1.322$ & $2.256$ && $0.046$ & $0.381$ & $1.377$ && $-0.517$ & $0.576$ & $1.291$ && $\mathbf{32.496}$ & $\mathbf{27.391}$ & $34.923$ && $\mathbf{0.145}$ & $\mathbf{0.121}$ & $0.088$ && $0.383$ & $0.398$ & $0.374$ \\

\cmidrule{2-30}
&\texttt{ NoIW-Random }&& $0.354$ & $1.898$ & $3.547$ && $0.912$ & $1.273$ & $2.654$ && $0.048$ & $0.796$ & $2.263$ && $-0.558$ & $0.625$ & $0.893$ && $13.775$ & $10.708$ & $10.677$ && $0.098$ & $0.072$ & $0.041$ && $0.365$ & $0.368$ & $0.364$ \\
&\texttt{ NoIW-ShortestPath }&& $0.354$ & $1.898$ & $3.547$ && $0.005$ & $0.005$ & $0.005$ && $0.005$ & $0.005$ & $0.005$ && $0.349$ & $1.893$ & $3.542$ && $0.000$ & $0.000$ & $0.000$ && $0.581$ & $0.581$ & $0.581$ && $0.451$ & $0.451$ & $0.451$ \\
\bottomrule
\end{tabular}

}\\[3pt]
\caption{Evaluation of \eqa agents trained \textbf{without} inflection weighting on navigation and answering metrics for the \mpeqa v1 test set.}
\label{tab:no_iw}
\end{table*}

%% file: sections/supp/main_tab_cis.tex
\begin{table*}[t!]
\setlength\tabcolsep{3pt}
\renewcommand{\arraystretch}{1.2}
\resizebox{\textwidth}{!}{
\begin{tabular}{l l l l c c c c c c l c c c l c c c l c c c l c c c l c c c }
& &  \multicolumn{24}{c}{\textbf{Navigation}} &~~~ & \multicolumn{3}{c}{\textbf{QA}}\\
\cmidrule{4-26}\cmidrule{28-30}
&\textbf{Navigator}&& \multicolumn{3}{c}{$\mathbf{d_0}$ {\scriptsize(For reference)}} &
& \multicolumn{3}{c}{$\mathbf{d_T}$ {\scriptsize(Lower is better)}} &
& \multicolumn{3}{c}{$\mathbf{d_{min}}$ {\scriptsize(Lower is better)}} &
& \multicolumn{3}{c}{$\mathbf{d_{\Delta}}$ {\scriptsize(Higher is better)}} &
& \multicolumn{3}{c}{$\mathbf{\%_{collision}}$ {\scriptsize(Lower is better)}} &
& \multicolumn{3}{c}{$\mathbf{IoU_{T}}$ {\scriptsize(Higher is better)}} &
& \multicolumn{3}{c}{$\mathbf{Top-1}$ {\scriptsize(Higher is better)}}\\
& && \scriptsize$T_{-10}$ & \scriptsize$T_{-30}$ & \scriptsize$T_{-50}$ &
& \scriptsize$T_{-10}$ & \scriptsize$T_{-30}$ & \scriptsize$T_{-50}$ &
& \scriptsize$T_{-10}$ & \scriptsize$T_{-30}$ & \scriptsize$T_{-50}$ &
& \scriptsize$T_{-10}$ & \scriptsize$T_{-30}$ & \scriptsize$T_{-50}$ &
 & \scriptsize$T_{-10}$ & \scriptsize$T_{-30}$ & \scriptsize$T_{-50}$ &
 & \scriptsize$T_{-10}$ & \scriptsize$T_{-30}$ & \scriptsize$T_{-50}$ &
 & \scriptsize$T_{-10}$ & \scriptsize$T_{-30}$ & \scriptsize$T_{-50}$ \\
 \toprule
&\texttt{ NoIW-R }&& $0.354$ & $1.898$ & $3.547$ && $0.933~(0.896,~0.969)$ & $1.330~(1.290,~1.370)$ & $2.154~(2.109,~2.196)$ && $\mathbf{0.011}~(0.010,~0.013)$ & $0.346~(0.333,~0.359)$ & $1.397~(1.367,~1.426)$ && $-0.579~(-0.615,~-0.542)$ & $0.568~(0.524,~0.611)$ & $1.393~(1.347,~1.443)$ && $79.554~(79.043,~80.065)$ & $66.182~(65.604,~66.768)$ & $62.563~(61.911,~63.244)$ && $0.062~(0.056,~0.069)$ & $0.050~(0.043,~0.056)$ & $0.030~(0.026,~0.035)$ && $0.390~(0.376,~0.404)$ & $0.379~(0.364,~0.392)$ & $0.354~(0.340,~0.367)$ \\
&\texttt{ NoIW-R+Q }&& $0.354$ & $1.898$ & $3.547$ && $0.933~(0.896,~0.969)$ & $1.330~(1.290,~1.370)$ & $2.154~(2.109,~2.196)$ && $\mathbf{0.011}~(0.010,~0.013)$ & $0.346~(0.333,~0.359)$ & $1.397~(1.367,~1.426)$ && $-0.579~(-0.615,~-0.542)$ & $0.568~(0.524,~0.611)$ & $1.393~(1.347,~1.443)$ && $79.554~(79.043,~80.065)$ & $66.182~(65.604,~66.768)$ & $62.563~(61.911,~63.244)$ && $0.062~(0.056,~0.069)$ & $0.050~(0.043,~0.056)$ & $0.030~(0.026,~0.035)$ && $0.390~(0.376,~0.404)$ & $0.379~(0.364,~0.392)$ & $0.354~(0.340,~0.367)$ \\
\\ 
&\texttt{ NoIW-R+RGB }&& $0.354$ & $1.898$ & $3.547$ && $1.194~(1.152,~1.234)$ & $1.617~(1.568,~1.665)$ & $2.340~(2.294,~2.385)$ && $0.040~(0.037,~0.043)$ & $0.375~(0.360,~0.390)$ & $1.349~(1.319,~1.379)$ && $-0.840~(-0.881,~-0.797)$ & $0.281~(0.230,~0.333)$ & $1.207~(1.156,~1.258)$ && $59.959~(58.970,~60.958)$ & $51.460~(50.585,~52.356)$ & $48.425~(47.578,~49.301)$ && $0.077~(0.069,~0.084)$ & $0.058~(0.051,~0.064)$ & $0.031~(0.026,~0.035)$ && $0.395~(0.380,~0.409)$ & $0.396~(0.381,~0.409)$ & $0.372~(0.358,~0.386)$ \\
&\texttt{ NoIW-R+RGB+Q }&& $0.354$ & $1.898$ & $3.547$ && $1.407~(1.361,~1.453)$ & $1.740~(1.692,~1.789)$ & $2.521~(2.467,~2.573)$ && $0.034~(0.031,~0.037)$ & $0.340~(0.327,~0.354)$ & $1.332~(1.301,~1.363)$ && $-1.053~(-1.101,~-1.005)$ & $0.157~(0.106,~0.209)$ & $1.026~(0.969,~1.085)$ && $51.128~(50.100,~52.154)$ & $44.160~(43.263,~45.066)$ & $42.692~(41.744,~43.618)$ && $0.111~(0.101,~0.121)$ & $0.070~(0.061,~0.078)$ & $0.054~(0.047,~0.061)$ && $0.383~(0.369,~0.397)$ & $0.388~(0.374,~0.403)$ & $0.375~(0.361,~0.389)$ \\
\\ 
&\texttt{ NoIW-R+PC }&& $0.354$ & $1.898$ & $3.547$ && $1.428~(1.379,~1.478)$ & $1.754~(1.706,~1.800)$ & $2.352~(2.299,~2.402)$ && $0.021~(0.019,~0.022)$ & $\mathbf{0.320}~(0.308,~0.333)$ & $1.164~(1.135,~1.192)$ && $-1.074~(-1.123,~-1.023)$ & $0.144~(0.092,~0.193)$ & $1.195~(1.140,~1.253)$ && $50.148~(49.084,~51.218)$ & $41.612~(40.672,~42.570)$ & $42.203~(41.276,~43.117)$ && $0.070~(0.063,~0.077)$ & $0.067~(0.060,~0.074)$ & $0.047~(0.042,~0.053)$ && $0.356~(0.342,~0.369)$ & $0.394~(0.380,~0.408)$ & $0.375~(0.361,~0.389)$ \\
&\texttt{ NoIW-R+PC+Q }&& $0.354$ & $1.898$ & $3.547$ && $1.514~(1.466,~1.563)$ & $1.812~(1.763,~1.862)$ & $2.394~(2.344,~2.445)$ && $0.033~(0.030,~0.035)$ & $0.325~(0.312,~0.338)$ & $\mathbf{1.160}~(1.131,~1.189)$ && $-1.160~(-1.209,~-1.110)$ & $0.085~(0.034,~0.138)$ & $1.153~(1.096,~1.210)$ && $46.910~(45.833,~47.956)$ & $36.303~(35.341,~37.244)$ & $39.012~(38.074,~39.942)$ && $0.059~(0.052,~0.065)$ & $0.052~(0.045,~0.058)$ & $0.043~(0.038,~0.048)$ && $0.364~(0.350,~0.378)$ & $0.364~(0.350,~0.378)$ & $0.363~(0.349,~0.377)$ \\
\\ 
&\texttt{ NoIW-R+PC+RGB }&& $0.354$ & $1.898$ & $3.547$ && $1.547~(1.497,~1.594)$ & $1.791~(1.743,~1.838)$ & $2.336~(2.287,~2.385)$ && $0.020~(0.018,~0.022)$ & $0.322~(0.309,~0.335)$ & $1.211~(1.182,~1.241)$ && $-1.193~(-1.241,~-1.143)$ & $0.107~(0.055,~0.159)$ & $1.211~(1.158,~1.265)$ && $44.941~(43.856,~46.040)$ & $34.859~(33.871,~35.853)$ & $37.138~(36.190,~38.091)$ && $0.084~(0.076,~0.093)$ & $0.077~(0.068,~0.086)$ & $0.044~(0.039,~0.049)$ && $0.374~(0.360,~0.388)$ & $0.390~(0.376,~0.404)$ & $0.366~(0.352,~0.380)$ \\
&\texttt{ NoIW-R+PC+RGB+Q }&& $0.354$ & $1.898$ & $3.547$ && $1.539~(1.490,~1.587)$ & $1.843~(1.795,~1.890)$ & $2.420~(2.369,~2.470)$ && $0.032~(0.029,~0.035)$ & $0.323~(0.311,~0.337)$ & $1.170~(1.141,~1.200)$ && $-1.185~(-1.233,~-1.137)$ & $0.055~(0.004,~0.107)$ & $1.127~(1.072,~1.186)$ && $42.018~(40.933,~43.100)$ & $34.318~(33.370,~35.271)$ & $37.069~(36.127,~38.004)$ && $0.067~(0.060,~0.075)$ & $0.072~(0.064,~0.080)$ & $0.055~(0.048,~0.062)$ && $0.370~(0.356,~0.384)$ & $0.395~(0.381,~0.409)$ & $0.369~(0.355,~0.382)$ \\

\cmidrule{2-30}
&\texttt{ NoIW-M }&& $0.354$ & $1.898$ & $3.547$ && $\mathbf{0.366}~(0.354,~0.379)$ & $\mathbf{0.830}~(0.811,~0.849)$ & $1.833~(1.800,~1.867)$ && $0.090~(0.085,~0.095)$ & $0.505~(0.490,~0.521)$ & $1.460~(1.432,~1.489)$ && $\mathbf{-0.012}~(-0.026,~0.001)$ & $\mathbf{1.068}~(1.042,~1.092)$ & $1.714~(1.674,~1.754)$ && $6.903~(6.508,~7.289)$ & $10.989~(10.516,~11.464)$ & $23.250~(22.632,~23.882)$ && $0.128~(0.117,~0.138)$ & $0.091~(0.083,~0.099)$ & $0.081~(0.072,~0.089)$ && $0.365~(0.351,~0.379)$ & $0.375~(0.361,~0.389)$ & $0.363~(0.349,~0.376)$ \\
&\texttt{ NoIW-M+Q }&& $0.354$ & $1.898$ & $3.547$ && $0.508~(0.491,~0.524)$ & $0.933~(0.912,~0.955)$ & $1.920~(1.884,~1.955)$ && $0.052~(0.048,~0.056)$ & $0.426~(0.412,~0.441)$ & $1.421~(1.392,~1.450)$ && $-0.154~(-0.171,~-0.136)$ & $0.965~(0.937,~0.992)$ & $1.627~(1.586,~1.669)$ && $16.268~(15.645,~16.890)$ & $19.808~(19.165,~20.458)$ & $32.856~(32.135,~33.588)$ && $0.147~(0.135,~0.157)$ & $0.109~(0.099,~0.119)$ & $0.068~(0.061,~0.075)$ && $0.391~(0.376,~0.405)$ & $0.395~(0.380,~0.409)$ & $0.376~(0.363,~0.390)$ \\
\\ 
&\texttt{ NoIW-M+RGB }&& $0.354$ & $1.898$ & $3.547$ && $0.637~(0.611,~0.663)$ & $1.157~(1.125,~1.187)$ & $2.177~(2.129,~2.224)$ && $0.099~(0.093,~0.105)$ & $0.538~(0.521,~0.556)$ & $1.479~(1.447,~1.510)$ && $-0.283~(-0.310,~-0.257)$ & $0.741~(0.707,~0.775)$ & $1.370~(1.319,~1.422)$ && $12.582~(11.849,~13.341)$ & $15.130~(14.404,~15.853)$ & $26.179~(25.319,~27.038)$ && $0.188~(0.174,~0.201)$ & $0.136~(0.125,~0.148)$ & $0.075~(0.067,~0.082)$ && $0.397~(0.384,~0.411)$ & $0.403~(0.388,~0.417)$ & $0.384~(0.370,~0.397)$ \\
&\texttt{ NoIW-M+RGB+Q }&& $0.354$ & $1.898$ & $3.547$ && $0.707~(0.681,~0.733)$ & $1.171~(1.140,~1.202)$ & $2.194~(2.146,~2.240)$ && $0.071~(0.066,~0.076)$ & $0.423~(0.408,~0.437)$ & $1.386~(1.355,~1.416)$ && $-0.353~(-0.380,~-0.327)$ & $0.727~(0.692,~0.762)$ & $1.353~(1.301,~1.406)$ && $14.212~(13.490,~14.925)$ & $15.908~(15.219,~16.604)$ & $25.578~(24.771,~26.392)$ && $0.189~(0.175,~0.203)$ & $0.141~(0.129,~0.153)$ & $0.083~(0.075,~0.091)$ && $\mathbf{0.407}~(0.392,~0.421)$ & $0.394~(0.380,~0.408)$ & $0.384~(0.370,~0.398)$ \\
\\ 
&\texttt{ NoIW-M+PC }&& $0.354$ & $1.898$ & $3.547$ && $0.494~(0.475,~0.512)$ & $1.020~(0.993,~1.047)$ & $1.817~(1.774,~1.859)$ && $0.098~(0.092,~0.103)$ & $0.484~(0.468,~0.500)$ & $1.236~(1.207,~1.265)$ && $-0.140~(-0.159,~-0.120)$ & $0.878~(0.846,~0.910)$ & $1.730~(1.683,~1.778)$ && $6.647~(6.073,~7.217)$ & $9.169~(8.522,~9.804)$ & $18.319~(17.510,~19.148)$ && $0.163~(0.152,~0.174)$ & $0.114~(0.104,~0.123)$ & $0.083~(0.076,~0.090)$ && $0.396~(0.381,~0.410)$ & $\mathbf{0.411}~(0.396,~0.425)$ & $\mathbf{0.390}~(0.376,~0.404)$ \\
&\texttt{ NoIW-M+PC+Q }&& $0.354$ & $1.898$ & $3.547$ && $0.502~(0.483,~0.521)$ & $1.030~(1.002,~1.059)$ & $1.910~(1.864,~1.954)$ && $0.081~(0.076,~0.086)$ & $0.497~(0.481,~0.514)$ & $1.272~(1.242,~1.302)$ && $-0.148~(-0.168,~-0.128)$ & $0.868~(0.836,~0.900)$ & $1.637~(1.588,~1.688)$ && $5.584~(5.105,~6.066)$ & $\mathbf{8.833}~(8.203,~9.454)$ & $\mathbf{15.783}~(15.007,~16.548)$ && $0.184~(0.171,~0.196)$ & $0.158~(0.146,~0.170)$ & $\mathbf{0.118}~(0.107,~0.128)$ && $0.382~(0.368,~0.396)$ & $0.387~(0.372,~0.401)$ & $0.374~(0.360,~0.388)$ \\
\\ 
&\texttt{ NoIW-M+PC+RGB }&& $0.354$ & $1.898$ & $3.547$ && $0.461~(0.443,~0.478)$ & $0.940~(0.915,~0.964)$ & $\mathbf{1.791}~(1.751,~1.831)$ && $0.103~(0.097,~0.109)$ & $0.513~(0.497,~0.529)$ & $1.269~(1.241,~1.297)$ && $-0.107~(-0.125,~-0.088)$ & $0.958~(0.928,~0.987)$ & $\mathbf{1.756}~(1.709,~1.802)$ && $\mathbf{4.957}~(4.519,~5.391)$ & $9.574~(8.973,~10.184)$ & $18.890~(18.105,~19.660)$ && $\mathbf{0.209}~(0.195,~0.222)$ & $\mathbf{0.179}~(0.166,~0.191)$ & $0.111~(0.102,~0.121)$ && $0.381~(0.367,~0.395)$ & $0.393~(0.378,~0.407)$ & $0.363~(0.350,~0.377)$ \\
&\texttt{ NoIW-M+PC+RGB+Q }&& $0.354$ & $1.898$ & $3.547$ && $0.574~(0.551,~0.598)$ & $1.044~(1.014,~1.074)$ & $1.898~(1.853,~1.941)$ && $0.083~(0.078,~0.088)$ & $0.431~(0.416,~0.446)$ & $1.203~(1.173,~1.232)$ && $-0.220~(-0.244,~-0.196)$ & $0.854~(0.819,~0.887)$ & $1.649~(1.600,~1.700)$ && $8.328~(7.687,~8.962)$ & $10.674~(10.000,~11.342)$ & $19.797~(18.973,~20.624)$ && $\mathbf{0.209}~(0.195,~0.222)$ & $0.148~(0.137,~0.160)$ & $0.112~(0.102,~0.123)$ && $0.389~(0.375,~0.404)$ & $0.390~(0.376,~0.404)$ & $0.373~(0.359,~0.387)$ \\

\cmidrule{2-30}
&\texttt{ NoIW-Random }&& $0.354$ & $1.898$ & $3.547$ && $0.912~(0.890,~0.935)$ & $1.273~(1.248,~1.298)$ & $2.654~(2.619,~2.686)$ && $0.048~(0.045,~0.051)$ & $0.796~(0.778,~0.815)$ & $2.263~(2.237,~2.289)$ && $-0.558~(-0.583,~-0.533)$ & $0.625~(0.597,~0.651)$ & $0.893~(0.861,~0.926)$ && $13.775~(13.476,~14.069)$ & $10.708~(10.431,~10.985)$ & $10.677~(10.408,~10.946)$ && $0.098~(0.088,~0.107)$ & $0.072~(0.064,~0.080)$ & $0.041~(0.035,~0.046)$ && $0.365~(0.352,~0.380)$ & $0.368~(0.354,~0.382)$ & $0.364~(0.350,~0.378)$ \\
&\texttt{ NoIW-ShortestPath }&& $0.354$ & $1.898$ & $3.547$ && $0.005~(0.004,~0.006)$ & $0.005~(0.004,~0.006)$ & $0.005~(0.004,~0.006)$ && $0.005~(0.004,~0.006)$ & $0.005~(0.004,~0.006)$ & $0.005~(0.004,~0.006)$ && $0.349~(0.340,~0.358)$ & $1.893~(1.876,~1.909)$ & $3.542~(3.522,~3.563)$ && $0.000~(0.000,~0.000)$ & $0.000~(0.000,~0.000)$ & $0.000~(0.000,~0.000)$ && $0.581~(0.567,~0.595)$ & $0.581~(0.567,~0.595)$ & $0.581~(0.567,~0.595)$ && $0.451~(0.437,~0.466)$ & $0.451~(0.437,~0.466)$ & $0.451~(0.437,~0.465)$ \\
\bottomrule
\end{tabular}

}\\[3pt]
\caption{\reftab{tab:main_results} (navigation results with inflection weightings) with 90\% bootstrap confidence intervals}
\label{tab:main_tab_cis}
\end{table*}

%% file: sections/supp/noiw_tab_cis.tex
\begin{table*}[t!]
\setlength\tabcolsep{3pt}
\renewcommand{\arraystretch}{1.2}
\resizebox{\textwidth}{!}{
\begin{tabular}{l l l l c c c c c c l c c c l c c c l c c c l c c c l c c c }
& &  \multicolumn{24}{c}{\textbf{Navigation}} &~~~ & \multicolumn{3}{c}{\textbf{QA}}\\
\cmidrule{4-26}\cmidrule{28-30}
&\textbf{Navigator}&& \multicolumn{3}{c}{$\mathbf{d_0}$ {\scriptsize(For reference)}} &
& \multicolumn{3}{c}{$\mathbf{d_T}$ {\scriptsize(Lower is better)}} &
& \multicolumn{3}{c}{$\mathbf{d_{min}}$ {\scriptsize(Lower is better)}} &
& \multicolumn{3}{c}{$\mathbf{d_{\Delta}}$ {\scriptsize(Higher is better)}} &
& \multicolumn{3}{c}{$\mathbf{\%_{collision}}$ {\scriptsize(Lower is better)}} &
& \multicolumn{3}{c}{$\mathbf{IoU_{T}}$ {\scriptsize(Higher is better)}} &
& \multicolumn{3}{c}{$\mathbf{Top-1}$ {\scriptsize(Higher is better)}}\\
& && \scriptsize$T_{-10}$ & \scriptsize$T_{-30}$ & \scriptsize$T_{-50}$ &
& \scriptsize$T_{-10}$ & \scriptsize$T_{-30}$ & \scriptsize$T_{-50}$ &
& \scriptsize$T_{-10}$ & \scriptsize$T_{-30}$ & \scriptsize$T_{-50}$ &
& \scriptsize$T_{-10}$ & \scriptsize$T_{-30}$ & \scriptsize$T_{-50}$ &
 & \scriptsize$T_{-10}$ & \scriptsize$T_{-30}$ & \scriptsize$T_{-50}$ &
 & \scriptsize$T_{-10}$ & \scriptsize$T_{-30}$ & \scriptsize$T_{-50}$ &
 & \scriptsize$T_{-10}$ & \scriptsize$T_{-30}$ & \scriptsize$T_{-50}$ \\
 \toprule
&\texttt{ NoIW-R }&& $0.354$ & $1.898$ & $3.547$ && $0.933~(0.896,~0.969)$ & $1.330~(1.290,~1.370)$ & $\mathbf{2.154}~(2.109,~2.196)$ && $\mathbf{0.011}~(0.010,~0.013)$ & $0.346~(0.333,~0.359)$ & $1.397~(1.367,~1.426)$ && $-0.579~(-0.615,~-0.542)$ & $0.568~(0.524,~0.611)$ & $\mathbf{1.393}~(1.347,~1.443)$ && $79.554~(79.043,~80.065)$ & $66.182~(65.604,~66.768)$ & $62.563~(61.911,~63.244)$ && $0.062~(0.056,~0.069)$ & $0.050~(0.043,~0.056)$ & $0.030~(0.026,~0.035)$ && $0.390~(0.376,~0.404)$ & $0.379~(0.364,~0.392)$ & $0.354~(0.340,~0.367)$ \\
&\texttt{ NoIW-R+Q }&& $0.354$ & $1.898$ & $3.547$ && $0.933~(0.896,~0.969)$ & $1.330~(1.290,~1.370)$ & $\mathbf{2.154}~(2.109,~2.196)$ && $\mathbf{0.011}~(0.010,~0.013)$ & $0.346~(0.333,~0.359)$ & $1.397~(1.367,~1.426)$ && $-0.579~(-0.615,~-0.542)$ & $0.568~(0.524,~0.611)$ & $\mathbf{1.393}~(1.347,~1.443)$ && $79.554~(79.043,~80.065)$ & $66.182~(65.604,~66.768)$ & $62.563~(61.911,~63.244)$ && $0.062~(0.056,~0.069)$ & $0.050~(0.043,~0.056)$ & $0.030~(0.026,~0.035)$ && $0.390~(0.376,~0.404)$ & $0.379~(0.364,~0.392)$ & $0.354~(0.340,~0.367)$ \\
\\ 
&\texttt{ NoIW-R+RGB }&& $0.354$ & $1.898$ & $3.547$ && $1.419~(1.372,~1.468)$ & $1.713~(1.666,~1.759)$ & $2.528~(2.475,~2.580)$ && $0.041~(0.037,~0.044)$ & $0.404~(0.389,~0.419)$ & $1.417~(1.385,~1.449)$ && $-1.065~(-1.114,~-1.016)$ & $0.185~(0.135,~0.235)$ & $1.019~(0.963,~1.077)$ && $56.718~(55.722,~57.732)$ & $50.376~(49.506,~51.248)$ & $45.866~(44.964,~46.762)$ && $0.086~(0.077,~0.094)$ & $0.049~(0.043,~0.055)$ & $0.035~(0.030,~0.040)$ && $0.382~(0.368,~0.396)$ & $0.386~(0.372,~0.400)$ & $0.360~(0.346,~0.373)$ \\
&\texttt{ NoIW-R+RGB+Q }&& $0.354$ & $1.898$ & $3.547$ && $1.405~(1.353,~1.454)$ & $1.829~(1.775,~1.882)$ & $2.658~(2.599,~2.716)$ && $0.051~(0.047,~0.055)$ & $0.455~(0.438,~0.472)$ & $1.463~(1.430,~1.496)$ && $-1.051~(-1.101,~-0.999)$ & $0.069~(0.012,~0.125)$ & $0.889~(0.827,~0.953)$ && $43.226~(42.106,~44.334)$ & $38.538~(37.559,~39.510)$ & $36.172~(35.216,~37.109)$ && $0.075~(0.066,~0.083)$ & $0.060~(0.053,~0.067)$ & $0.054~(0.047,~0.061)$ && $0.384~(0.369,~0.398)$ & $0.390~(0.376,~0.404)$ & $0.373~(0.359,~0.387)$ \\
\\ 
&\texttt{ NoIW-R+PC }&& $0.354$ & $1.898$ & $3.547$ && $1.385~(1.338,~1.430)$ & $1.662~(1.616,~1.708)$ & $2.483~(2.428,~2.535)$ && $0.026~(0.023,~0.028)$ & $\mathbf{0.343}~(0.329,~0.358)$ & $1.294~(1.263,~1.324)$ && $-1.031~(-1.078,~-0.983)$ & $0.236~(0.185,~0.286)$ & $1.064~(1.005,~1.123)$ && $43.067~(41.951,~44.169)$ & $34.100~(33.167,~35.046)$ & $37.078~(36.097,~38.050)$ && $0.074~(0.066,~0.082)$ & $0.061~(0.054,~0.068)$ & $0.043~(0.038,~0.048)$ && $0.375~(0.361,~0.389)$ & $0.398~(0.384,~0.412)$ & $0.376~(0.362,~0.390)$ \\
&\texttt{ NoIW-R+PC+Q }&& $0.354$ & $1.898$ & $3.547$ && $1.515~(1.464,~1.564)$ & $1.858~(1.806,~1.910)$ & $2.646~(2.589,~2.700)$ && $0.038~(0.035,~0.042)$ & $0.394~(0.378,~0.410)$ & $1.333~(1.302,~1.365)$ && $-1.161~(-1.212,~-1.108)$ & $0.040~(-0.016,~0.096)$ & $0.901~(0.842,~0.962)$ && $37.669~(36.572,~38.772)$ & $31.714~(30.788,~32.640)$ & $\mathbf{33.563}~(32.594,~34.548)$ && $0.078~(0.070,~0.085)$ & $0.059~(0.052,~0.066)$ & $0.054~(0.047,~0.060)$ && $0.372~(0.359,~0.387)$ & $\mathbf{0.399}~(0.385,~0.413)$ & $0.378~(0.364,~0.392)$ \\
\\ 
&\texttt{ NoIW-R+PC+RGB }&& $0.354$ & $1.898$ & $3.547$ && $1.462~(1.415,~1.508)$ & $1.759~(1.712,~1.806)$ & $2.523~(2.468,~2.576)$ && $0.025~(0.022,~0.027)$ & $0.347~(0.332,~0.362)$ & $\mathbf{1.285}~(1.254,~1.316)$ && $-1.108~(-1.155,~-1.060)$ & $0.139~(0.087,~0.189)$ & $1.024~(0.966,~1.084)$ && $53.785~(52.783,~54.794)$ & $41.955~(41.038,~42.882)$ & $40.293~(39.360,~41.224)$ && $0.067~(0.059,~0.075)$ & $0.052~(0.045,~0.058)$ & $0.042~(0.036,~0.047)$ && $0.384~(0.370,~0.398)$ & $0.389~(0.375,~0.403)$ & $0.369~(0.356,~0.383)$ \\
&\texttt{ NoIW-R+PC+RGB+Q }&& $0.354$ & $1.898$ & $3.547$ && $1.297~(1.251,~1.342)$ & $1.704~(1.654,~1.753)$ & $2.543~(2.489,~2.595)$ && $0.030~(0.028,~0.033)$ & $0.425~(0.408,~0.441)$ & $1.417~(1.385,~1.449)$ && $-0.943~(-0.989,~-0.897)$ & $0.194~(0.142,~0.247)$ & $1.004~(0.947,~1.063)$ && $47.506~(46.410,~48.588)$ & $39.554~(38.569,~40.565)$ & $36.242~(35.263,~37.210)$ && $0.069~(0.062,~0.076)$ & $0.062~(0.055,~0.069)$ & $0.046~(0.040,~0.052)$ && $0.368~(0.354,~0.382)$ & $0.375~(0.360,~0.389)$ & $0.353~(0.340,~0.367)$ \\

\cmidrule{2-30}
&\texttt{ NoIW-M }&& $0.354$ & $1.898$ & $3.547$ && $0.933~(0.896,~0.969)$ & $1.330~(1.290,~1.370)$ & $2.186~(2.141,~2.229)$ && $\mathbf{0.011}~(0.010,~0.013)$ & $0.346~(0.333,~0.359)$ & $1.430~(1.400,~1.460)$ && $-0.579~(-0.615,~-0.542)$ & $0.568~(0.524,~0.611)$ & $1.361~(1.314,~1.410)$ && $79.554~(79.043,~80.065)$ & $66.182~(65.604,~66.768)$ & $62.818~(62.155,~63.487)$ && $0.062~(0.056,~0.069)$ & $0.050~(0.043,~0.056)$ & $0.029~(0.025,~0.033)$ && $0.390~(0.376,~0.404)$ & $0.379~(0.364,~0.392)$ & $0.356~(0.341,~0.369)$ \\
&\texttt{ NoIW-M+Q }&& $0.354$ & $1.898$ & $3.547$ && $0.933~(0.896,~0.969)$ & $1.330~(1.290,~1.370)$ & $2.202~(2.156,~2.244)$ && $\mathbf{0.011}~(0.010,~0.013)$ & $0.346~(0.333,~0.359)$ & $1.445~(1.415,~1.475)$ && $-0.579~(-0.615,~-0.542)$ & $0.568~(0.524,~0.611)$ & $1.345~(1.299,~1.395)$ && $79.554~(79.043,~80.065)$ & $66.182~(65.604,~66.768)$ & $62.931~(62.273,~63.603)$ && $0.062~(0.056,~0.069)$ & $0.050~(0.043,~0.056)$ & $0.029~(0.025,~0.033)$ && $0.390~(0.376,~0.404)$ & $0.379~(0.364,~0.392)$ & $0.354~(0.340,~0.368)$ \\
\\ 
&\texttt{ NoIW-M+RGB }&& $0.354$ & $1.898$ & $3.547$ && $0.902~(0.866,~0.937)$ & $1.396~(1.355,~1.436)$ & $2.512~(2.456,~2.566)$ && $0.024~(0.021,~0.026)$ & $0.400~(0.385,~0.415)$ & $1.608~(1.574,~1.643)$ && $-0.548~(-0.584,~-0.511)$ & $0.502~(0.458,~0.547)$ & $1.035~(0.980,~1.093)$ && $67.347~(66.413,~68.309)$ & $59.661~(58.861,~60.466)$ & $59.378~(58.660,~60.133)$ && $0.080~(0.072,~0.088)$ & $0.061~(0.054,~0.068)$ & $0.033~(0.028,~0.038)$ && $\mathbf{0.406}~(0.392,~0.420)$ & $0.384~(0.370,~0.398)$ & $0.353~(0.339,~0.366)$ \\
&\texttt{ NoIW-M+RGB+Q }&& $0.354$ & $1.898$ & $3.547$ && $0.911~(0.874,~0.948)$ & $1.394~(1.351,~1.435)$ & $2.573~(2.515,~2.630)$ && $0.024~(0.021,~0.026)$ & $0.410~(0.394,~0.425)$ & $1.644~(1.609,~1.678)$ && $-0.557~(-0.594,~-0.519)$ & $0.504~(0.460,~0.549)$ & $0.974~(0.917,~1.034)$ && $66.198~(65.281,~67.131)$ & $59.317~(58.523,~60.122)$ & $58.941~(58.212,~59.682)$ && $0.094~(0.086,~0.103)$ & $0.064~(0.057,~0.071)$ & $0.030~(0.026,~0.035)$ && $0.390~(0.376,~0.404)$ & $0.380~(0.365,~0.393)$ & $0.356~(0.343,~0.370)$ \\
\\ 
&\texttt{ NoIW-M+PC }&& $0.354$ & $1.898$ & $3.547$ && $0.811~(0.780,~0.841)$ & $1.245~(1.209,~1.282)$ & $2.244~(2.194,~2.292)$ && $0.034~(0.031,~0.037)$ & $0.370~(0.357,~0.385)$ & $1.414~(1.383,~1.447)$ && $-0.457~(-0.488,~-0.425)$ & $0.652~(0.612,~0.693)$ & $1.303~(1.251,~1.357)$ && $37.865~(36.674,~39.002)$ & $30.964~(29.981,~31.969)$ & $38.521~(37.532,~39.517)$ && $0.113~(0.103,~0.122)$ & $0.114~(0.104,~0.123)$ & $\mathbf{0.091}~(0.081,~0.100)$ && $0.386~(0.372,~0.401)$ & $\mathbf{0.399}~(0.385,~0.413)$ & $0.376~(0.362,~0.390)$ \\
&\texttt{ NoIW-M+PC+Q }&& $0.354$ & $1.898$ & $3.547$ && $\mathbf{0.790}~(0.760,~0.820)$ & $\mathbf{1.233}~(1.197,~1.269)$ & $2.213~(2.163,~2.262)$ && $0.038~(0.034,~0.041)$ & $0.379~(0.366,~0.393)$ & $1.416~(1.385,~1.448)$ && $\mathbf{-0.436}~(-0.466,~-0.404)$ & $\mathbf{0.665}~(0.626,~0.704)$ & $1.334~(1.282,~1.387)$ && $36.215~(35.059,~37.351)$ & $31.431~(30.439,~32.446)$ & $38.911~(37.929,~39.887)$ && $0.115~(0.106,~0.124)$ & $0.099~(0.090,~0.108)$ & $0.076~(0.068,~0.083)$ && $0.393~(0.379,~0.407)$ & $\mathbf{0.399}~(0.385,~0.413)$ & $\mathbf{0.379}~(0.364,~0.392)$ \\
\\ 
&\texttt{ NoIW-M+PC+RGB }&& $0.354$ & $1.898$ & $3.547$ && $0.929~(0.894,~0.965)$ & $1.405~(1.362,~1.448)$ & $2.435~(2.378,~2.490)$ && $0.016~(0.014,~0.018)$ & $0.375~(0.360,~0.389)$ & $1.526~(1.494,~1.558)$ && $-0.575~(-0.610,~-0.539)$ & $0.492~(0.446,~0.539)$ & $1.112~(1.055,~1.171)$ && $61.658~(60.620,~62.668)$ & $51.595~(50.659,~52.541)$ & $52.093~(51.216,~52.969)$ && $0.100~(0.092,~0.109)$ & $0.075~(0.068,~0.082)$ & $0.041~(0.036,~0.046)$ && $0.388~(0.374,~0.402)$ & $0.385~(0.371,~0.400)$ & $0.353~(0.339,~0.366)$ \\
&\texttt{ NoIW-M+PC+RGB+Q }&& $0.354$ & $1.898$ & $3.547$ && $0.871~(0.837,~0.905)$ & $1.322~(1.283,~1.362)$ & $2.256~(2.203,~2.308)$ && $0.046~(0.043,~0.050)$ & $0.381~(0.367,~0.396)$ & $1.377~(1.345,~1.410)$ && $-0.517~(-0.551,~-0.482)$ & $0.576~(0.532,~0.620)$ & $1.291~(1.236,~1.349)$ && $\mathbf{32.496}~(31.347,~33.629)$ & $\mathbf{27.391}~(26.415,~28.397)$ & $34.923~(33.916,~35.928)$ && $\mathbf{0.145}~(0.133,~0.157)$ & $\mathbf{0.121}~(0.111,~0.131)$ & $0.088~(0.079,~0.098)$ && $0.383~(0.369,~0.398)$ & $0.398~(0.384,~0.413)$ & $0.374~(0.360,~0.388)$ \\

\cmidrule{2-30}
&\texttt{ NoIW-Random }&& $0.354$ & $1.898$ & $3.547$ && $0.912~(0.890,~0.935)$ & $1.273~(1.248,~1.298)$ & $2.654~(2.619,~2.686)$ && $0.048~(0.045,~0.051)$ & $0.796~(0.778,~0.815)$ & $2.263~(2.237,~2.289)$ && $-0.558~(-0.583,~-0.533)$ & $0.625~(0.597,~0.651)$ & $0.893~(0.861,~0.926)$ && $13.775~(13.476,~14.069)$ & $10.708~(10.431,~10.985)$ & $10.677~(10.408,~10.946)$ && $0.098~(0.088,~0.107)$ & $0.072~(0.064,~0.080)$ & $0.041~(0.035,~0.046)$ && $0.365~(0.352,~0.380)$ & $0.368~(0.354,~0.382)$ & $0.364~(0.350,~0.378)$ \\
&\texttt{ NoIW-ShortestPath }&& $0.354$ & $1.898$ & $3.547$ && $0.005~(0.004,~0.006)$ & $0.005~(0.004,~0.006)$ & $0.005~(0.004,~0.006)$ && $0.005~(0.004,~0.006)$ & $0.005~(0.004,~0.006)$ & $0.005~(0.004,~0.006)$ && $0.349~(0.340,~0.358)$ & $1.893~(1.876,~1.909)$ & $3.542~(3.522,~3.563)$ && $0.000~(0.000,~0.000)$ & $0.000~(0.000,~0.000)$ & $0.000~(0.000,~0.000)$ && $0.581~(0.567,~0.595)$ & $0.581~(0.567,~0.595)$ & $0.581~(0.567,~0.595)$ && $0.451~(0.437,~0.466)$ & $0.451~(0.437,~0.466)$ & $0.451~(0.437,~0.465)$ \\
\bottomrule
\end{tabular}

}\\[3pt]
\caption{\reftab{tab:no_iw_cis} (navigation results \textbf{without} inflection weightings) with 90\% bootstrap confidence intervals}
\label{tab:no_iw_cis}
\end{table*}